\documentclass{article}

\usepackage{microtype}
\usepackage{subfigure}
\usepackage{booktabs} 
\usepackage{subcaption} 
\usepackage{wrapfig}

\usepackage{hyperref}

\usepackage[accepted]{icml2024}

\usepackage{amsmath}
\usepackage{amssymb}
\usepackage{mathtools}
\usepackage{amsthm}

\usepackage{mathrsfs}
\usepackage{commath}
\usepackage[capitalize,noabbrev]{cleveref}

\theoremstyle{plain}
\newtheorem{theorem}{Theorem}[section]

\theoremstyle{definition}
\newtheorem{definition}[theorem]{Definition}

\theoremstyle{remark}

\usepackage[textsize=tiny]{todonotes}

\newcommand{\OPT}{\operatorname{OPT}}

\newcommand{\Nat}{\text{Na}}
\newcommand{\Cat}{\text{Ca}}
\newcommand{\Kt}{\text{K}}
\newcommand{\Mt}{\text{M}}
\newcommand{\Lt}{\text{L}}
\newcommand{\leakt}{\text{leak}}
\newcommand{\maxt}{\text{max}}

\DeclareMathOperator*{\argmax}{arg\,max}

\icmltitlerunning{Diffusion Tempering Improves Parameter Estimation for ODEs}

\begin{document}

\twocolumn[
\icmltitle{Diffusion Tempering Improves Parameter Estimation with Probabilistic Integrators for Ordinary Differential Equations}


\icmlsetsymbol{equal}{*}

\begin{icmlauthorlist}
\icmlauthor{Jonas Beck}{HertieAI,TueUni}
\icmlauthor{Nathanael Bosch}{TueUni}
\icmlauthor{Michael Deistler}{TueUni}
\icmlauthor{Kyra L. Kadhim}{HertieAI,TueUni}
\icmlauthor{Jakob H. Macke}{TueUni,MPI}
\icmlauthor{Philipp Hennig}{TueUni}
\icmlauthor{Philipp Berens}{HertieAI,TueUni}
\end{icmlauthorlist}

\icmlaffiliation{TueUni}{AI Center, University of Tübingen, Tübingen, Germany}
\icmlaffiliation{HertieAI}{Hertie Institute for AI in Brain Health, University of Tübingen, Tübingen, Germany}
\icmlaffiliation{MPI}{Max Planck Institute for Intelligent Systems, Tübingen, Germany}

\icmlcorrespondingauthor{Jonas Beck}{jonas.beck@uni-tuebingen.de}

\icmlkeywords{Machine Learning, Hodgkin--Huxley, Probabilistic Numerics, ODE, Optimization, Parameter Estimation}

\vskip 0.3in
]


\printAffiliationsAndNotice{}  

\begin{abstract}
Ordinary differential equations (ODEs) are widely used to describe dynamical systems in science, but identifying parameters that explain experimental measurements is challenging. In particular, although ODEs are differentiable and would allow for gradient-based parameter optimization, the nonlinear dynamics of ODEs often lead to many local minima and extreme sensitivity to initial conditions. We therefore propose diffusion tempering, a novel regularization technique for probabilistic numerical methods which improves convergence of gradient-based parameter optimization in ODEs. By iteratively reducing a noise parameter of the probabilistic integrator, the proposed method converges more reliably to the true parameters. We demonstrate that our method is effective for dynamical systems of different complexity and show that it obtains reliable parameter estimates for a Hodgkin--Huxley model with a practically relevant number of parameters.
\end{abstract}

\section{Introduction}\label{sec:intro}
Ordinary differential equations (ODEs) are ubiquitous across science, as they often provide accurate models for physical processes and mechanisms underlying dynamical systems. ODEs can, for example, be used to model the motion of a pendulum, the dynamics of predator and prey populations \cite{Hethcote_2000} or the action potentials of neurons \citep{Hodgkin_Huxley_1952}. Numerical algorithms for the (approximate) solution of ODEs are well established \cite{Hairer_Norsett_Wanner_1987}, but identifying parameters of the ODE such that it matches observations can be challenging \citep{Prinz_Billimoria_Marder_2003, Kravtsov_2020}. 
To overcome this, multiple methods have been developed, such as grid \cite{Prinz_Billimoria_Marder_2003} or random searches \cite{Taylor_Goaillard_Marder_2009}, simulated annealing \cite{Kirkpatrick_Gelatt_Vecchi_1983}, genetic algorithms \cite{Ben-Shalom_Aviv_Razon_Korngreen_2012}, Bayesian methods (e.g., approximate Bayesian computation \cite{Marjoram_Molitor_Plagnol_Tavaré_2003}, Markov-Chain Monte-Carlo \cite{Neal_1993}, or simulation-based inference \cite{Cranmer_Brehmer_Louppe_2020}). But, these methods can often require large simulation budgets \cite{Cranmer_Brehmer_Louppe_2020}.

By incorporating gradient information, gradient descent provides the potential to be vastly more simulation efficient and, as demonstrated by Neural ODEs (NODEs), has the potential to scale to millions of parameters \cite{Chen_Rubanova_Bettencourt_Duvenaud_2018}. Unfortunately, for many real-world ODEs, gradient-based optimization is challenging: ODEs often have highly nonlinear dynamics, several local minima, and are sensitive to initial conditions \cite{Cao_Wang_Xu_2011, Dass_Lee_Lee_Park_2017}. In fact, even for models with few parameters, gradient-based optimization can get stuck in local minima, leading to poor performance compared to competing gradient-free methods such as genetic algorithms \cite{Hazelden_Liu_Shlizerman_Shea-Brown_2023}.

Probabilistic numerical integrators have also been proposed for parameter inference in ODEs
\citep{Tronarp_Bosch_Hennig_2022}.
Unlike `traditional' non-probabilistic ODE solvers which usually return only a single (potentially very crude) ODE solution, probabilistic numerical methods return a posterior distribution over the ODE solution, accounting for numerical uncertainty 
\citep{Hennig_Osborne_Kersting_2022}.
Parameters can then be estimated by maximizing their likelihood, which thanks to automatic differentiation is amenable to first order methods. 
This approach has been termed ``Physics-Enhanced Regression for Initial Value Problems'', or Fenrir for short.

Here, we develop \emph{diffusion tempering} for probabilistic integrators to improve gradient-based parameter inference in ODEs. The technique is based on previous observations that Fenrir can effectively `smooth out' the loss surface of ODEs \cite{Tronarp_Bosch_Hennig_2022}. We therefore propose to solve consecutive optimization problems using the Fenrir likelihood. We start with a very smooth loss surface, which yields poor fits to data, but lets the optimizer avoid local minima. By successively solving less and less smooth problems informed by previous parameter estimates, we more reliably converge in the global optimum. This enables gradient-based maximum likelihood estimation for models with a practically relevant number of parameters such as \citet{Pospischil_2008}.

We first demonstrate that our method is robust to local minima for a simple pendulum. We then show that it produces more reliable parameter estimates than both classical least-squares regression and the original Fenrir method by \citet{Tronarp_Bosch_Hennig_2022} for several models of growing complexity, even in regimes in which gradient-based parameter inference is challenging.

\section{Parameter inference in ODEs}

Consider an initial value problem (IVP), given by an ODE
\begin{align}\label{eq:ivp}
    \dot{y}_\theta(t) = f_\theta \! \left( y_\theta(t), t \right), \qquad t \in [0, T],
\end{align}
with vector field $f_\theta$ with parameters \(\theta\), and initial value $y_\theta(0) = y_0$.
In this paper, we are concerned with estimating the true ODE parameters \(\theta\) from a set of noisy observations of the solution \(y_\theta(t)\), of the form
\begin{align}\label{eq:observation_model}
  u_i = u(t_i) = H y_\theta(t_i) + \epsilon_i, \qquad i = 1,\dots,N,
\end{align}
where \(H\) is a measurement matrix and \(\epsilon_i \sim \mathcal{N}(0, R)\) is Gaussian noise with covariance \(R\).
We denote the set of observations by \(\mathcal{D} = \{u_i\}_{i=1}^N\), and the set of observation times by \(\mathbb{T}_\mathcal{D} = \{t_i\}_{i=1}^N\).

For a given parameter \(\theta\), the true ODE solution \(y_\theta(t)\) is uniquely defined and thus the true marginal likelihood is 
\begin{align}\label{eq:exact_marginal_likelihood}
  \mathcal{M}(\theta) = \mathcal{L}_\mathcal{D}(y_\theta) = \prod_{i=1}^N \mathcal{N}(u_i; H y_\theta(t_i), R),
\end{align}
where \(\mathcal{L}_\mathcal{D}\) denotes the likelihood functional.
Then, the parameter \(\theta\) can be estimated from the data \(\mathcal{D}\) by maximizing the marginal likelihood \(\mathcal{M}(\theta)\), that is
\(
  {\hat{\theta}_\text{MLE} = \arg\max_\theta \mathcal{M}(\theta).}
\)
Unfortunately, the true solution \(y_\theta\) is not generally known and cannot be computed analytically, and thus the true marginal likelihood is intractable.

In the following, we discuss two approaches to make inference tractable by approximating the marginal likelihood.
First, note that we can rewrite 
\(\mathcal{M}(\theta)\)
as an integral 
\begin{align}\label{eq:exact_marginal_likelihood2}
  \mathcal{M}(\theta) = \int \mathcal{L}_\mathcal{D}(y) \delta(y - y_\theta) \dif y.
\end{align}
This clarifies that the intractable part of the marginal likelihood is the Dirac measure \(\delta(y - y_\theta)\).
Thus, a natural approach to approximate the marginal likelihood is to replace the true solution distribution 
by a suitable approximation.

\subsection{Classic Numerical Integration}\label{sec:rk}
A standard approach to parameter inference in IVPs approximates the true IVP solution \(y_\theta(t)\) with a numerical solution \(\hat{y}_\theta(t)\), obtained from a classic numerical IVP solver such as the well-known Runge--Kutta (RK) method \cite{Hairer_Norsett_Wanner_1987}.
The marginal likelihood then becomes
\begin{subequations}
\begin{align}\label{eq:rk:approx_marginal_likelihood}
  \mathcal{M}(\theta) \approx
  \widehat{\mathcal{M}}_\text{RK}(\theta)
  :=& \int \mathcal{L}_\mathcal{D}(y) \delta(y - \hat{y}_\theta) \dif y, \\
  =& \prod_{i=1}^N \mathcal{N}(u_i; H \hat{y}_\theta(t_i), R).
\end{align}
\end{subequations}
Maximizing $\widehat{\mathcal{M}}_\text{RK}(\theta)$ is equivalent to minimizing
the mean squared error
\begin{align}\label{eq:rk:loss}
  L_\text{RK}(\theta) = \frac{1}{N} \sum_{i=1}^N \left\| H \hat{y}_\theta(t_i) - u_i \right\|_2^2.
\end{align}
This approach is also known as non-linear least-squares regression \cite{Bard_1974}.
 
\subsection{Probabilistic Numerical Integration}\label{sec:fenrir}
Probabilistic numerical methods reformulate numerical problems as probabilistic inference \cite{Hennig_Osborne_Kersting_2022}.
In the context of IVPs, probabilistic numerical ODE solvers make the time-discretization of numerical methods explicitly part of their problem statement, and aim to compute a posterior distribution of the form 
\begin{align}\label{eq:pn:posterior}
    p \left( y_\theta(t) ~\big|~
    y_\theta(0) = y_0, 
    \{ \dot{y}_\theta(t) = f_\theta\!\left( y_\theta(t), t \right) \}_{t \in \mathbb{T}_\text{PN}} 
    \right),
\end{align}
where \(\mathbb{T}_\text{PN}\) is the chosen time-discretization.
Assuming fixed \(f\), \(y_0\), and \(\mathbb{T}_\text{PN}\), we denote this posterior more compactly with \( p_\text{PN} ( y(t) \mid \theta) \).
We call this object, and approximations thereof, a \emph{probabilistic numerical ODE solution} 

This object then presents itself as a natural candidate to approximate the true solution distribution \(\delta(y - y_\theta)\).
We obtain the PN-approximated marginal likelihood 
\begin{align}\label{eq:pn:approx_marginal_likelihood}
  \widehat{\mathcal{M}}_\text{PN}(\theta)
  &= \int \mathcal{L}_\mathcal{D}(y) \ p_\text{PN} ( y(t) \mid \theta ) \dif y.
\end{align}
The remaining question is how to compute this quantity.
In the following, we consider probabilistic ODE solvers based on Bayesian filtering and smoothing, which have been shown to be a particularly efficient class of methods for probabilistic numerical simulation
\cite{Tronarp_Särkkä_Hennig_2021,tronarp18,kersting18,schober16},
and we review the PN-approximated marginal likelihood by
\citet{Tronarp_Bosch_Hennig_2022}
which builds the base of our proposed approach.

\subsubsection{Probabilistic Numerical IVP Solvers}\label{sec:pn_solvers}
Filtering-based probabilistic numerical ODE solvers formulate the probabilistic numerical ODE solution 
(\cref{eq:pn:posterior}) 
as a Bayesian state estimation problem, 
described by a Gauss--Markov prior given by
\begin{subequations}\label{eq:ivp2gmr}
\begin{align}
x(0) &\sim \mathcal{N}\!\left( \mu_0, \Sigma_0 \right), \\
x(t_n) \mid x(t_{n-1}) &\sim \mathcal{N}\!\left( \Phi_n x(t_{n-1}), \kappa^2 Q_n \right), \\
y^{(m)}(t_n) &= E_m^\intercal x(t_n), \quad m = 0, 1, \ldots, q,
\intertext{and a likelihood and data model%
, or \emph{information operator}, z \cite{Tronarp_Särkkä_Hennig_2021,Cockayne_Oates_Sullivan_Girolami_2019},
of the form
}
z(t_n) \mid x(t_n) &\sim \delta\!\left( E_1 x(t_n) - f_\theta\!\left(E_0 x(t_n), t\right) \right), \\
z(t_n)  &\coloneqq 0,
\end{align}
\end{subequations}
which maps solutions of the initial value problem to the zero function on the chosen grid $\mathbb{T}_\mathcal{D}$.

Here, the state \(x(t)\) models the solution \(y(t)\) of the IVP together with its $q$ first derivatives, 
and in the following we consider \(q\)-times integrated Wiener process priors for which the transition matrices \(\Phi_n\) and \(Q_n\) are known in closed form \citep{kersting18}.
$E_m$ are selection matrices for the $m$th derivative.
The initial mean \(\mu_0\) is chosen to match the initial condition of the given IVP exactly, and the initial covariance \(\Sigma_0\) is set to zero
\citep{kraemer20_stabl_implem_probab_ode_solver}.
Finally, $\kappa$ is a prior hyperparameter which controls the uncertainty of the prior distribution, known as the \emph{diffusion}.

\Cref{eq:ivp2gmr} is a well known problem in Bayesian filtering and smoothing, 
and its solution can be approximated efficiently with extended Kalman filtering and smoothing
\cite{Särkkä_2013}.
For a given ODE parameter \(\theta\) and diffusion hyperparameter \(\kappa\), we obtain a Gauss--Markov posterior distribution 
\(\hat{p}_\text{PN} ( y(t) \mid \theta, \kappa )\).

\subsubsection{PN-approximated Marginal Likelihood}
By inserting the PN posterior 
\(\hat{p}_\text{PN} ( y(t) \mid \theta, \kappa )\),
into the marginal likelihood, we obtain the marginal likelihood model by 
\citet{Tronarp_Bosch_Hennig_2022}
of the form
\begin{align}\label{eq:pn:fenrir_marginal_likelihood}
  \widehat{\mathcal{M}}_\text{FN}(\theta, \kappa)
  &= \int \mathcal{L}_\mathcal{D}(y) \ \hat{p}_\text{PN} ( y(t) \mid \theta, \kappa ) \dif y.
\end{align}
This quantity can again be computed efficiently with Kalman filtering:
Since the posterior distribution 
\(\hat{p}_\text{PN}\)
has known backward transition densities
of the form
\citep[Proposition 3.1]{Tronarp_Bosch_Hennig_2022}
\begin{align}\label{eq:pn:backward_transitions}
\begin{split}
  \hat{p}_\text{PN}\! \left( 
  x(t_{n-1}) \mid x(t_n) 
  \right) = 
  \mathcal{N} \big( G^\theta_n x(t_n) + \zeta^\theta_n, \kappa^2 P^\theta_n \big),
\end{split}
\end{align}
\(\widehat{\mathcal{M}}_\text{FN}(\theta, \kappa)\)
can be computed by running a Kalman filter backwards in time on the state-space model
\begin{subequations}\label{eq:fenrir}
\begin{align}
x(t_N) &\sim \mathcal{N} \big( x(t_N); \xi_\theta(t_N), \kappa^2 \Lambda_\theta(t_N) \big), \\
x(t_{n-1}) \mid x(t_n) &\sim \mathcal{N} \big( G^\theta_n x(t_n) + \zeta^\theta_n, \kappa^2 P^\theta_n \big), \\
u(t_n) \mid x(t_n) &\sim \mathcal{N} \big( H E_0 x(t_n), R_\theta \big), \quad t_n \in \mathbb{T}_{\mathsf{D}}, 
\end{align}
\end{subequations}
via the prediction error decomposition \cite{Schweppe_1965}.
The quantities \(G^\theta_n, \zeta^\theta_n, P^\theta_n\) can all be computed during the forward-pass of the probabilistic ODE solver; for the full details refer to \citet{Tronarp_Bosch_Hennig_2022}.

Finally, 
\citet{Tronarp_Bosch_Hennig_2022}
propose to jointly optimize
\(\widehat{\mathcal{M}}_\text{FN}(\theta, \kappa)\)
for both the ODE parameters of interest \(\theta\) and the diffusion hyperparameter \(\kappa\), that is
\begin{align}\label{eq:pn:fenrir_max_marginal_likelihood}
\hat{\theta}, \hat{\kappa} = 
\argmax_{\theta,\kappa}
\widehat{\mathcal{M}}_\text{FN}\!\left(\theta, \kappa\right).
\end{align}
Then \(\hat\theta\) is returned as the maximum likelihood estimate.
We refer to this ODE parameter inference method as Fenrir.

\begin{figure}[t]\centering
\centerline{\includegraphics[width=\columnwidth]{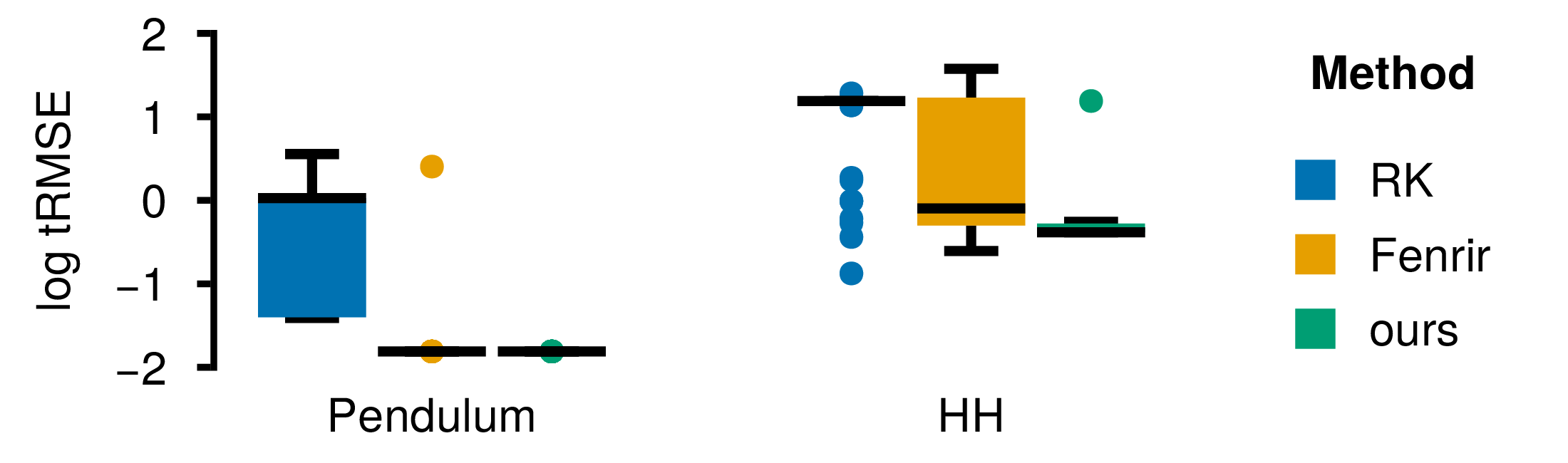}}
\caption{For a pendulum model, Fenrir produces parameter estimates with much lower trajectory mean squared errors (\cref{def:trmse}) than RK least-squares regression. However, for the more complex HH model, the tRMSE is very similar for both RK and Fenrir. Our proposed method is able to produce much better estimates for both problems.}
\label{fig:figure1}
\end{figure}

\subsection{Shortcomings}
In order to ensure that an optimizer has converged at the global as opposed to a local optimum, in practice, the optimization has to be run multiple times with different initial conditions. A reliable optimizer is therefore highly desirable since a higher reliability also means less restarts to obtain a good parameter estimate. This in turn can make optimizing more complex models feasible, since we can afford many more restarts at the same cost. 

\citet{Tronarp_Bosch_Hennig_2022} show that learning $\theta$ and $\kappa$ jointly leads to more reliable parameter estimates for a pendulum compared to RK least-squares regression (\cref{fig:figure1}) for which the optimization often converges in local minima, i.e. the constant zero function. However, a pendulum is a fairly low dimensional problem. When we evaluated Fenrir for more complex ODEs such as the HH model, we found that this gain in reliability was vastly smaller in systems with more challenging likelihood functions. In these cases, many restarts are needed to reach the global optimum, which makes Fenrir equally unfeasible as RK-based least squares for gradient descent-based parameter estimation in HH models with a practically relevant number of parameters.

\section{Method}
In the following, we attribute the observed shortcomings of Fenrir to the way the diffusion hyperparameter $\kappa$ was treated and we propose an alternative method which enables effective parameter inference for more complicated loss functions and higher dimensional parameter spaces (\cref{fig:figure1}).

\subsection{Diffusion as Regularization}\label{sec:diffusion_regularization}
To gain a better understanding of the mechanisms that lead to better optimization performance in Fenrir, we studied the effect of the diffusion hyperparameter in more detail.

\begin{figure}[t]\centering
\centerline{\includegraphics[width=\columnwidth]{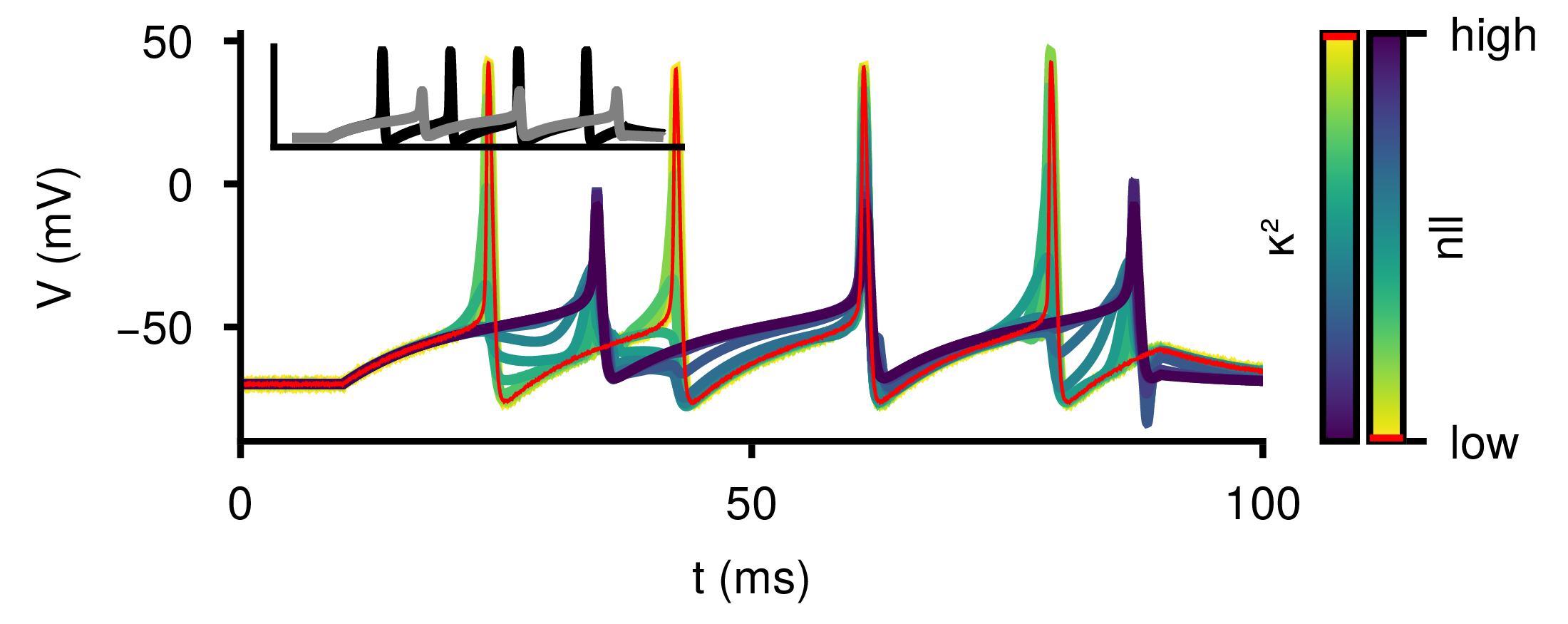}}
\caption{PN posterior means for a range of $\kappa$ for a HH model. The parameters that generated the observation (inset, black) are different from the parameters of the ODE (inset, grey). For low $\kappa$ the mean closely adheres to the ODE solution, while for high values, it fits the observation much better. The maximum likelihood $\kappa$ is highlighted in red.}
\label{fig:figure2}
\end{figure}

\begin{algorithm}[tb]
   \caption{Diffusion tempering}
   \label{alg:diffusion_tempering}
\begin{algorithmic}
   \STATE {\bfseries Input:} initial parameter $\theta_0$, tempering schedule $\mathcal{T}$, number of iterations $m$, Optimizer $\OPT$, Fenrir likelihood $\widehat{\mathcal{M}}(\theta, \kappa)$
   \FOR{$i=0$ {\bfseries to} $m-1$}
   \STATE $\kappa_i = \mathcal{T}(i)$ 
   \STATE $\theta_{i+1} = \OPT\!\left(\widehat{\mathcal{M}}_\text{PN}, (\theta_{i}, \kappa_i)\right)$
   \ENDFOR
   \STATE $\theta_{est} = \theta_{m}$
   \STATE {\bfseries Return: $\theta_{est}$}
\end{algorithmic}
\end{algorithm}

We noticed that for a given set of model parameters $\theta$, the loss was lower for high values of $\kappa$ than for low values (\cref{fig:figure2}). Furthermore, when $\kappa$ was set high, the mean of the PN posterior interpolated the observed data points, while if set low, it approximated the IVP solution much better. This implies for this particular parameter set, that the maximum likelihood estimate of the PN posterior (\cref{eq:pn:fenrir_max_marginal_likelihood}) ``favors'' a good interpolation of the data over a good approximation of the IVP solution. This is an issue for the success of parameter estimation, since we are trying to identify an IVP that reproduces the observed data well, as opposed to finding just a good interpolant of the data.

Why does this happen? Since $\kappa$ scales the gain of the diffusion, it also determines the uncertainty of the ODE solution for a given $\theta$. A high diffusion leads to a less concentrated prior for the subsequent Gauss--Markov regression. For a broad prior, many observations have a similar likelihood, hence a PN posterior whose mean interpolates the data scores favorably (\cref{fig:figure2}). For a low diffusion, the variance of the prior is very tightly concentrated around the ODE solution, meaning only data that closely matches the trajectory will score a high likelihood, in turn leading to interpolation of the ODE solution (\cref{fig:figure2}). 

Based on these observations, we propose to re-interpret the diffusion parameter in Fenrir as a optimization hyperparameter which determines how the data should be ``weighted'' with respect to the solution of the ODE: Fixing $\kappa$ regularizes the solution of the IVP on the observed data. In particular, we show in \cref{fig:figure1} and argue in the following that tempering the diffusion leads to improved convergence.

\begin{figure*}[t]
\centering
\includegraphics[width=\textwidth]{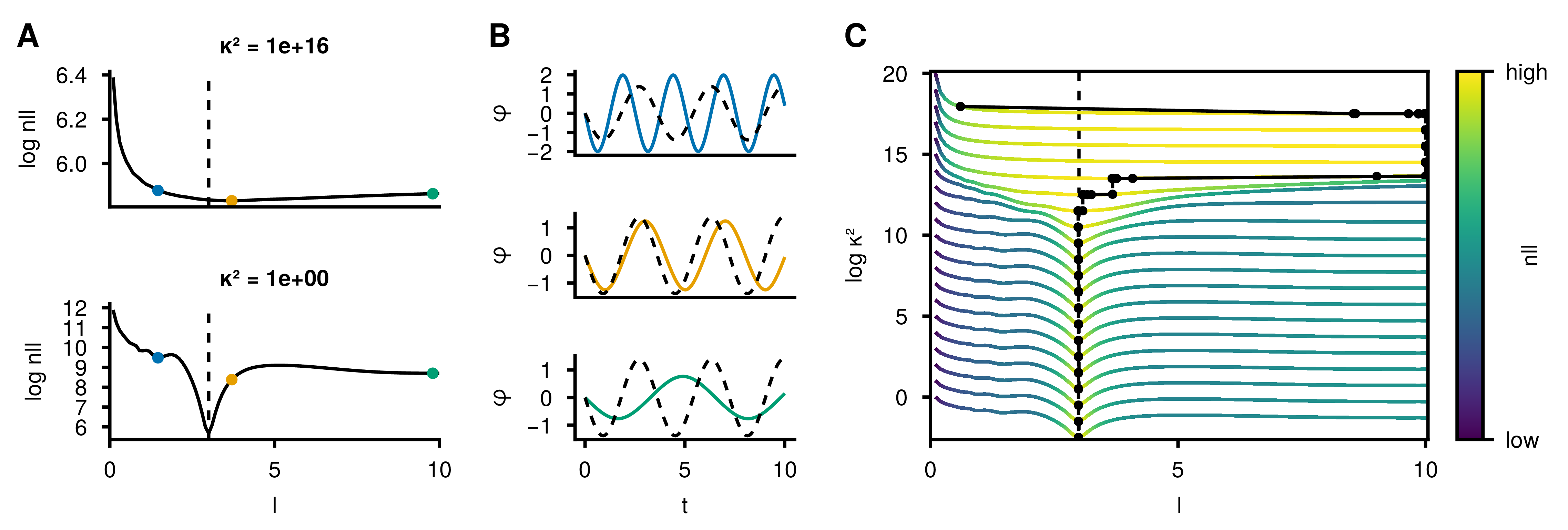}
\caption{The effect of diffusion tempering on the marginal likelihood $\mathcal{M}(\theta, \kappa)$ of a pendulum with parameter $l$. \textbf{A} The negative log likelihood (nll) for a high $\kappa$ is very smooth with only one shallow global optimum (yellow). The likelihood for low $\kappa$ has one sharp global minimum at the true parameters (dashed), but also other local minima (blue, green). \textbf{B} IVP solutions for the different local optima marked in A and the true solution (dashed). \textbf{C} Diffusion tempering for an exemplary optimization run (black): Optimization starts with a high $\kappa$ (top) on a very smooth loss landscape. $\kappa$ is then progressively lowered, revealing a shallow global optimum near the true parameters $(\kappa=10^{16})$. Further lowering $\kappa$ finally reveals a sharp global optimum and the optimization converges correctly. By starting with the parameters of the previous optimization, diffusion tempering ensures that the optimization closes in on the correct global optimum.}
\label{fig:figure3}
\end{figure*}

\subsection{Diffusion tempering}\label{sec:diffusion_tempering}
Above we argued why $\kappa$ should not be optimized jointly with $\theta$, but treated as a hyperparameter during optimization, since it can trade-off confidence in the data vs.~the IVP solution (Fig. \ref{fig:figure2}).
Accordingly, we propose to put $\kappa$ on a decreasing schedule.
Starting with a high $\kappa$, Fenrir does not have much confidence in the accuracy of its physics-informed prior and therefore most ODE parameters yield a decent likelihood for the observed data. By steadily lowering the diffusion, we decrease the uncertainty of the prior, which means there are less ways for an ODE solution to fit the data. Finally, when the prior has minimal uncertainty presumably only the ``true" ODE fits the data.

We call this procedure diffusion tempering (\cref{alg:diffusion_tempering}). We start by defining the number of iterations $m$ and a tempering schedule $\mathcal{T}$, which can be a list of $\kappa$ or any function that takes a positive integer and returns a value for the diffusion hyperparameter. We set an initial parameter $\theta_0$, i.e. by drawing it from a pre-specified distribution and set $\widehat{\mathcal{M}}_\text{PN}(\theta_0, \kappa=\mathcal{T}(0))$. We  run an optimizer of choice until convergence to obtain a new parameter estimate. Then, we obtain the next $\kappa$ from $\mathcal{T}$ and repeat the optimization with the updated estimate of $\theta$, returning the final $\theta$ as our parameter estimate.
In our case, we initialized the parameters uniformly distributed and opted for a tempering schedule of $\mathcal{T}(i, \kappa_0) = 10^{(\kappa_0 - i)}$, $\kappa_0=20$ with a schedule length of $21$. This is equivalent to linearly decreasing \(\log_{10}(\kappa)\) from $20$ to $0$.

\section{Experiments}
We first investigate the effect of diffusion tempering for a pendulum with a single parameter, and show even in this comparatively simple case that it converges more reliably around the true parameters. We then demonstrate that diffusion tempering also improves Fenrir's convergence for complex ODE models such as the HH model \citep{Hodgkin_Huxley_1952}. Finally, we show that diffusion tempering enables parameter estimation for HH models in parameter regimes where learned diffusions and least-squares methods stop working altogether. 

\subsection{Exploration of a simple case: The 1D Pendulum}\label{sec:pendulum_results}
We first demonstrate our algorithm for a pendulum. While this is a relatively simple problem by construction, it is already challenging for classical simulation-based methods such as Runge–Kutta least-squares approaches \cite{Bard_1974}, which tend to fail for high frequencies and are sensitive to initialization \cite{Benson_1979}. Furthermore, with only a single free parameter, the pendulum length $l$, the effect of tempered optimization can be illustrated well (\cref{fig:figure3}). Details on the full dynamics and the parameterisation are provided in \cref{sec:PD_appendix}.

We evaluated the Fenrir marginal likelihood at different pendulum lengths for a high diffusion constant. This confirmed our theory from \cref{sec:diffusion_regularization} that a high $\kappa$ indeed leads to similar likelihoods for most parameters (\cref{fig:figure3}A). Additionally, with increasing length \(l\) the loss also increased, which discourages the zero-function (\(l \to \infty\)) which would otherwise be a local optimum. However, while the global optimum was roughly in the right place, it was not yet located at the true $l$. This was in stark contrast to when we evaluated the marginal likelihood for a low diffusion. Here the global optimum correctly identified the true $l$, but the loss landscape had additional local optima (\cref{fig:figure3}B). When we repeated this process for every $\kappa$ that our tempering schedule visited, we observed that solving separate optimization problems at subsequently lower values of $\kappa$ exploited the smooth nature of the loss landscape to skip over later forming local optima. While earlier minima might not have been accurate, they provided a better initialization for following optimizations at lower $\kappa$ (\cref{fig:figure3}C). Diffusion tempering first interpolated the observed data, but later put more and more weight on a good ODE solution. This avoided local minima that provided a likelihood that explained neither the ODE or the data well. For the pendulum, diffusion tempering converged as often as Fenrir but three times as often as RK least-squares regression in a ball around the true parameters (\cref{tab:benchmarks}, PD).

\begin{figure*}[t]
\centering
\includegraphics[width=\textwidth]{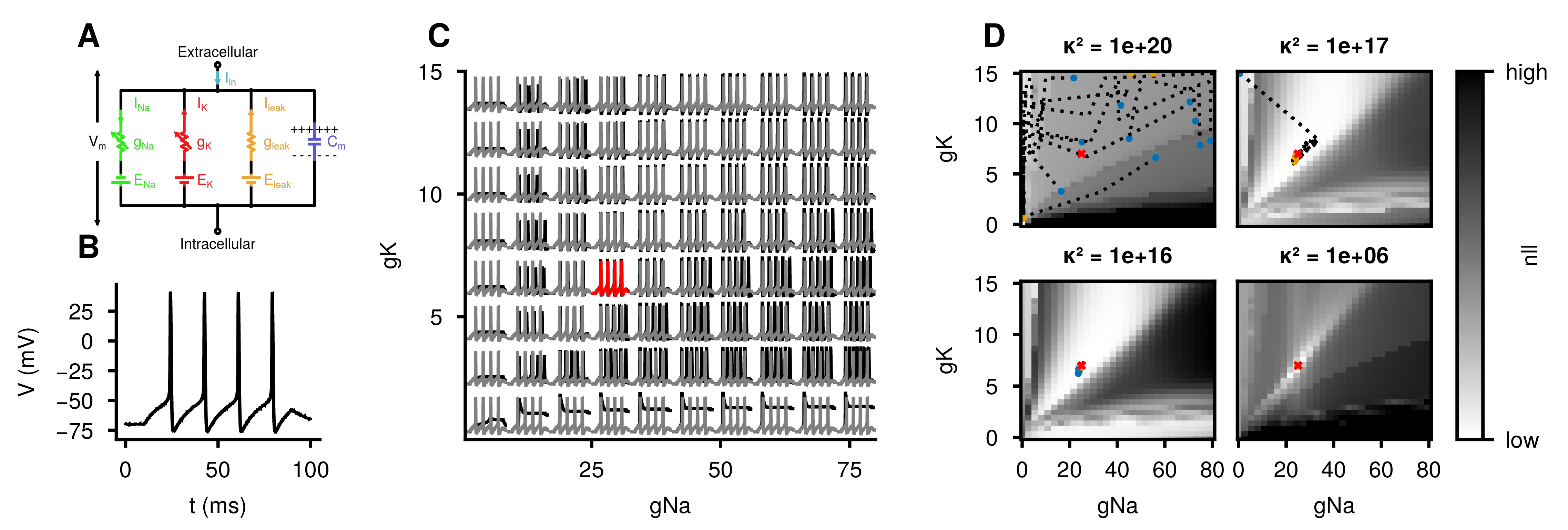}
\caption{Parameter estimation for the sodium $g_{Na}$ and potassium $g_{K}$ conductance of a HH model. \textbf{A} Circuit modeling components of a neuronal cell as electrical elements. The lipid bilayer acts as a capacitor ($C_m$). Ion channels are represented by resistors. Sodium and potassium conductances are voltage dependent ($g_{Na}$,$g_{K}$), and the leak conductance is constant ($g_{leak}$). The electrochemical gradients driving the flow of ions can be represented as voltage sources ($E_{Na}$,$E_{K}$,$E_{leak}$). \textbf{B} Noisy observation of the membrane voltage. \textbf{C} Solutions of the ODE for different combinations of parameters (black) compared to the true parameters (grey). The true parameters are highlighted in red. \textbf{D} Parameter optimization during four different stages of diffusion tempering for a random subset of initializations. Optimization trajectories (dotted) in each likelihood landscape are shown from start (blue) to convergence (orange). Very high loss values were clipped for better visual clarity. Plots for the full schedule are provided in \cref{fig:appendix_figure4a}, with corresponding solutions in \cref{fig:appendix_figure4b}.}
\label{fig:figure4}
\end{figure*}

\subsection{Exploration of a complex case: The Hodgkin--Huxley Model} \label{sec:HH_results}
We next turned to a more complex system of ODEs, the Hodgkin--Huxley model for neural membrane biophysics \citep{Hodgkin_Huxley_1952}. It models the change in voltage across a cell's membrane $V(t)$ in response to an external input $I_{\text{in}}(t)$ as different ionic currents flowing through a circuit (\cref{fig:figure4}A). For the detailed equations and parameterisation see \cref{sec:HH_appendix}.

This model is an interesting case study because there is considerable scientific interest in identifying the conductances of these ionic currents from voltage recordings to inform neuron simulations. All-or-none action potentials (spikes) and steep voltage gradients make this challenging, which is why grid \cite{Prinz_Billimoria_Marder_2003} or random searches \cite{Taylor_Goaillard_Marder_2009}, genetic algorithms \cite{Druckmann_2007, Ben-Shalom_Aviv_Razon_Korngreen_2012, Achard_De_Schutter_2006} and simulation-based Inference \cite{Goncalves_2020, Lueckmann_2017} are common tools.

For our experiments, we generated observations from the HH model with a square wave depolarizing current of 210 pA between 10ms and 90 ms as stimulus. In the initial experiments, we only considered two free parameters, $g_{\text{Na}}$ and $g_{\text{K}}$, for which we generated data at 25 mS/cm$^2$ and 7 mS/cm$^2$, respectively. The initial voltage was chosen as $V(0)=-70$ mV and the membrane voltage was simulated for a time interval of 100 ms, leading to a spiking trace with four action potentials (\cref{fig:figure4}B). See \cref{sec:HH_appendix} for details on how we initialized the gating variables.

\begin{table*}[t]
\centering
\caption{Comparison of different methods, models and model sizes for parameter estimation.  LV: Lottka-Volterra, PD: Pendulum, $\text{HH}_x$ Hodgkin--Huxley with $x$ compartments, $\text{D}_\theta$: Number of parameters, $\text{N}_\theta$: The number of correctly identified parameters, CONV: correctly converged. \emph{In only a few cases the optimization diverged, which lead to $NaN$s in the loss and tRMSEs orders of magnitude above the mean. We excluded these runs from the statistics marked with a $^*$}.}
\vskip 0.15in
\begin{footnotesize}
\begin{sc}
\begin{tabular}{llllllll}
  \hline
  \textbf{ODE} & \textbf{$\text{D}_\theta$} & \textbf{ALG} & \textbf{ITER} & \textbf{pRMSE} & \textbf{CONV} & \textbf{tRMSE} & \textbf{$\text{N}_\theta$} \\
  \hline
  PD & 1 & Fenrir & $74.12 \pm 11.55$ & $0.01 \pm 0.08$ & 0.99 & $0.04 \pm 0.26$ & $0.99 \pm 0.10$ \\
  PD & 1 & RK & $\mathbf{34.11 \pm 18.45}$ & $1.42 \pm 1.09$ & 0.32 & $0.98 \pm 0.72$ & $0.32 \pm 0.47$ \\
  PD & 1 & ours & $303.89 \pm 12.43$ & $\mathbf{0.00 \pm 0.00}$ & $\mathbf{1.00}$ & $\mathbf{0.02 \pm 0.00}$ & $\mathbf{1.00 \pm 0.00}$ \\
  PD & 1 & ours+ & $92.21 \pm 0.95$ & $\mathbf{0.00 \pm 0.00}$ & $\mathbf{1.00}$ & $\mathbf{0.02 \pm 0.00}$ & $\mathbf{1.00 \pm 0.00}$ \\
  \hline
  LV & 2 & Fenrir & $\mathbf{60.68 \pm 36.73}$ & $1.41 \pm 1.01$ & 0.20 & $2.84 \pm 2.71$ & $0.40 \pm 0.80$ \\
  LV & 2 & RK & $97.15 \pm 146.95$ & $1.35 \pm 1.11$ & 0.24 & $2.35 \pm 1.32$ & $0.48 \pm 0.86$ \\
  LV & 2 & ours & $411.18 \pm 151.25$ & $0.41 \pm 0.82$ & 0.77 & $^*0.54 \pm 1.14$ & $1.54 \pm 0.85$ \\
  LV & 2 & ours+ & $132.43 \pm 31.01$ & $\mathbf{0.35 \pm 0.77}$ & $\mathbf{0.79}$ & $^*\mathbf{0.43 \pm 1.03}$ & $\mathbf{1.59 \pm 0.81}$ \\
  \hline
  LV & 4 & Fenrir & $\mathbf{112.45 \pm 47.24}$ & $1.31 \pm 0.94$ & 0.23 & $\mathbf{4.18 \pm 13.65}$ & $1.07 \pm 1.70$ \\
  LV & 4 & RK & $271.49 \pm 151.05$ & $1.09 \pm 0.76$ & 0.24 & $6.57 \pm 34.45$ & $0.97 \pm 1.71$ \\
  LV & 4 & ours & $727.15 \pm 482.93$ & $0.81 \pm 0.88$ & 0.43 & $^*10.41 \pm 69.37$ & $1.76 \pm 1.98$ \\
  LV & 4 & ours+ & $292.05 \pm 139.98$ & $\mathbf{0.74 \pm 0.82}$ & $\mathbf{0.44}$ & $^*13.77 \pm 79.08$ & $\mathbf{1.79 \pm 1.98}$ \\
  \hline
  $\text{HH}_1$ & 1 & Fenrir & $46.96 \pm 19.08$ & $0.38 \pm 0.67$ & 0.68 & $7.85 \pm 10.70$ & $0.68 \pm 0.47$ \\
  $\text{HH}_1$ & 1 & RK & $\mathbf{43.30 \pm 43.45}$ & $0.42 \pm 0.48$ & 0.57 & $7.54 \pm 8.26$ & $0.57 \pm 0.50$ \\
  $\text{HH}_1$ & 1 & ours & $382.08 \pm 32.19$ & $\mathbf{0.00 \pm 0.00}$ & $\mathbf{1.00}$ & $\mathbf{0.43 \pm 0.02}$ & $\mathbf{1.00 \pm 0.00}$ \\
  $\text{HH}_1$ & 1 & ours+ & $116.73 \pm 7.11$ & $\mathbf{0.00 \pm 0.00}$ & $\mathbf{1.00}$ & $0.43 \pm 0.11$ & $\mathbf{1.00 \pm 0.00}$ \\
  \hline
  $\text{HH}_1$ & 2 & Fenrir & $110.04 \pm 61.70$ & $0.20 \pm 0.37$ & 0.75 & $5.89 \pm 10.15$ & $1.53 \pm 0.83$ \\
  $\text{HH}_1$ & 2 & RK & $\mathbf{54.02 \pm 62.60}$ & $0.28 \pm 0.45$ & 0.72 & $4.88 \pm 7.58$ & $1.44 \pm 0.90$ \\
  $\text{HH}_1$ & 2 & ours & $696.22 \pm 78.10$ & $\mathbf{0.00 \pm 0.00}$ & $\mathbf{1.00}$ & $\mathbf{0.42 \pm 0.04}$ & $\mathbf{2.00 \pm 0.00}$ \\
  $\text{HH}_1$ & 2 & ours+ & $197.81 \pm 40.38$ & $0.04 \pm 0.19$ & 0.96 & $1.08 \pm 3.20$ & $1.92 \pm 0.39$ \\
  \hline
  $\text{HH}_1$ & 3 & Fenrir & $\mathbf{122.15 \pm 49.74}$ & $0.59 \pm 0.65$ & 0.51 & $9.31 \pm 9.63$ & $1.53 \pm 1.51$ \\
  $\text{HH}_1$ & 3 & RK & $223.55 \pm 117.31$ & $0.90 \pm 0.25$ & 0.03 & $14.33 \pm 4.01$ & $0.13 \pm 0.56$ \\
  $\text{HH}_1$ & 3 & ours & $676.33 \pm 150.72$ & $\mathbf{0.01 \pm 0.10}$ & $\mathbf{0.99}$ & $\mathbf{0.60 \pm 1.51}$ & $\mathbf{2.97 \pm 0.30}$ \\
  \hline
  $\text{HH}_1$ & 6 & Fenrir & $\mathbf{108.06 \pm 108.49}$ & $13.36 \pm 6.97$ & 0.00 & $26.21 \pm 7.38$ & $1.05 \pm 0.22$ \\
  $\text{HH}_1$ & 6 & RK & $210.26 \pm 120.86$ & $12.27 \pm 6.87$ & 0.00 & $16.80 \pm 3.81$ & $1.18 \pm 0.39$ \\
  $\text{HH}_1$ & 6 & ours & $2159.60 \pm 532.55$ & $\mathbf{10.36 \pm 7.72}$ & 0.00 & $^*\mathbf{15.20 \pm 5.41}$ & $\mathbf{1.21 \pm 0.46}$ \\
  \hline
  $\text{HH}_2$ & 4 & Fenrir & $286.54 \pm 205.37$ & $0.28 \pm 0.44$ & 0.68 & $12.00 \pm 17.37$ & $2.80 \pm 1.78$ \\
  $\text{HH}_2$ & 4 & RK & $\mathbf{136.50 \pm 200.20}$ & $0.43 \pm 0.56$ & 0.50 & $7.98 \pm 9.86$ & $2.08 \pm 1.81$ \\
  $\text{HH}_2$ & 4 & ours & $1492.03 \pm 335.17$ & $\mathbf{0.00 \pm 0.00}$ & $\mathbf{1.00}$ & $\mathbf{0.60 \pm 0.01}$ & $\mathbf{4.00 \pm 0.00}$ \\
  \hline
  $\text{HH}_2$ & 6 & Fenrir & $\mathbf{221.28 \pm 144.56}$ & $0.62 \pm 0.72$ & 0.50 & $13.01 \pm 13.10$ & $3.06 \pm 2.96$ \\
  $\text{HH}_2$ & 6 & RK & $390.34 \pm 195.85$ & $0.88 \pm 0.22$ & 0.00 & $19.36 \pm 5.54$ & $0.31 \pm 0.75$ \\
  $\text{HH}_2$ & 6 & ours & $1525.57 \pm 448.56$ & $\mathbf{0.12 \pm 0.32}$ & $\mathbf{0.88}$ & $\mathbf{3.01 \pm 6.70}$ & $\mathbf{5.28 \pm 1.96}$ \\
  \hline
\end{tabular}
\label{tab:benchmarks}
\end{sc}
\end{footnotesize}
\end{table*}

Even with just two free parameters, the HH model displayed highly complex patterns of activity ranging from non-spiking solutions in regimes of low sodium or potassium conductance to spike trains of different amplitudes, frequencies or phase (\cref{fig:figure4}C). Additionally, small changes in the parameter space could sometimes lead to dramatic changes in the solution, like additional spikes appearing.

To explore this and the effect on the inference more systematically, we computed the loss landscape of the model with respect to observed data for different diffusion parameters (\cref{fig:figure4}D). We observed a similar behavior as in \cref{sec:pendulum_results} for the pendulum. For a high diffusion parameter, the loss was very smooth and sloped only in one direction, leading the optimization runs to converge at the non-spiking response simply interpolating the baseline of the action potentials (\cref{fig:figure4}D top, left). This likely happened because the action potentials were narrow, consisting only of a few measurements each. Hence, their effect on the loss was not strong enough to make these solutions prohibitively costly, as the high diffusion parameter made the inference focus on data interpolation. For lower diffusion parameters, a large basin around the true parameters appeared, which the parameter estimates started to converge to (\cref{fig:figure4}D top, right). For very low diffusion values, the basin then morphed into a much sharper global optimum around the true parameters which attracted the estimates that were previously collecting in the basin (\cref{fig:figure4}D bottom, left).

\subsection{Systematic performance benchmarking against alternative methods}\label{sec:benchmark_results}
To demonstrate the effect of tempering the diffusion more systematically, we compared to both learning and fixing $\kappa$ in the following. More specifically we studied:  
(i) non-linear least-squares regression with Runge--Kutta (RK),
(ii) Fenrir with learned diffusion as proposed by \citet{Tronarp_Bosch_Hennig_2022},
and (iii) Fenrir with a range of fixed diffusion parameters: 
We explore a low, a high and the best $\kappa$ according to a grid search over all values visited during tempering (for details see \cref{fig:appendix_figure5}) as well as the maximum likelihood $\kappa$ at the true parameters (optimal).

We considered an optimization run as correctly converged if the final parameter estimate deviated at most 5\% from the true parameters (for details see \cref{sec:metrics}).

\begin{figure}[t]
\centering
\includegraphics[width=\columnwidth]{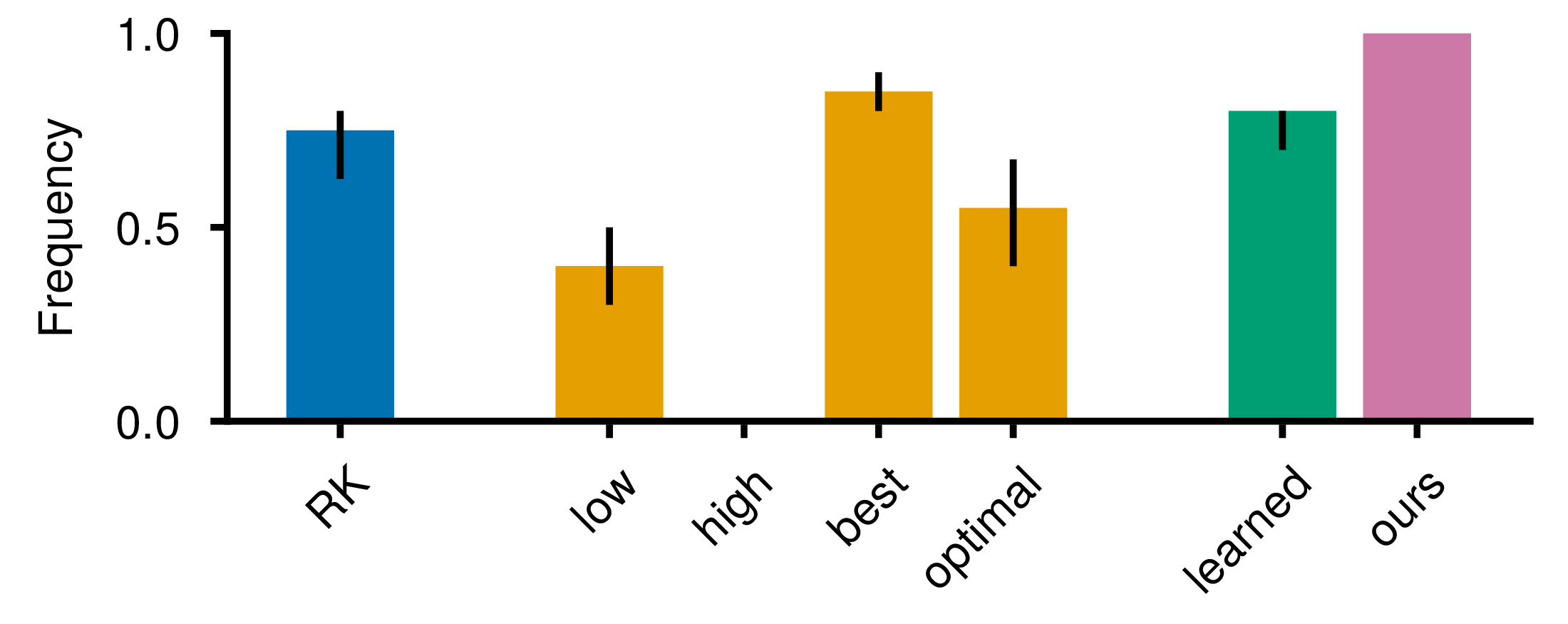}
\caption{Convergence for a two parameter HH model: Diffusion tempering converges more reliably than than non-linear least-squares regression using a RK solver and Fenrir with a learned and the best single $\kappa$. Histogram shows the medians with black bars indicating quartiles for 100 runs split into groups of 10.}
\label{fig:figure5}
\end{figure}

\begin{figure*}[t]
\centering
\includegraphics[width=\textwidth]{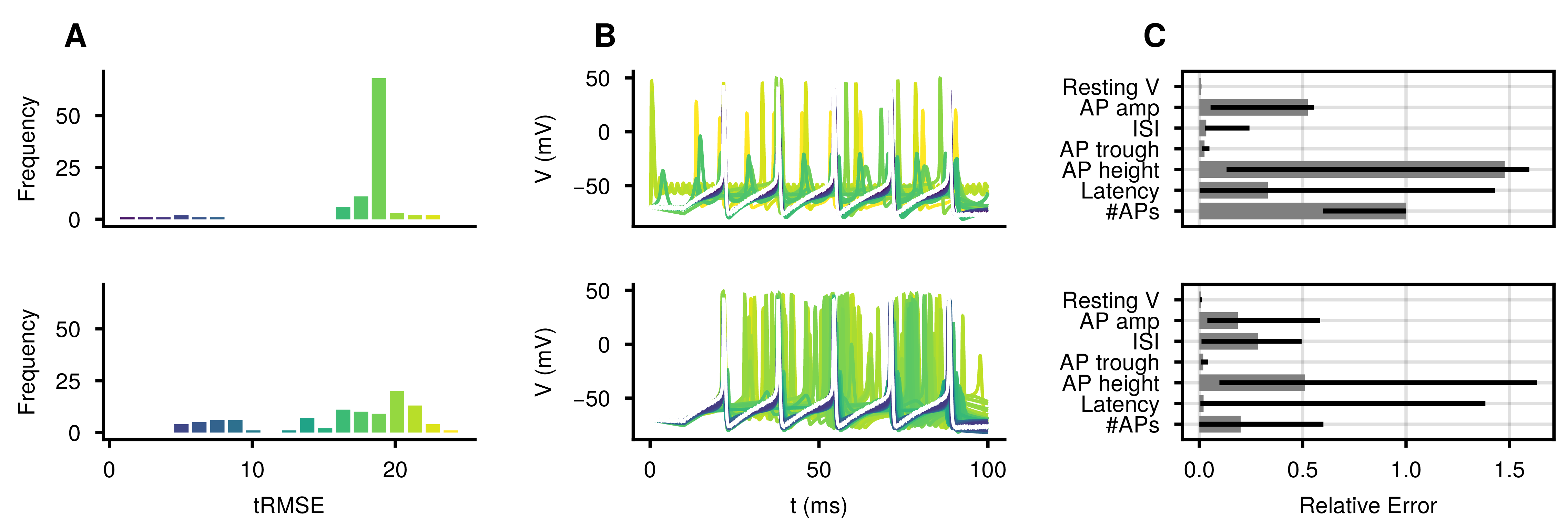}
\caption{Parameter inference for a six parameter HH model for 100 initializations of RK (top) and tempered Fenrir (bottom). \textbf{A} Distribution of tRMSEs. The average tRMSE for Fenrir is lower than for RK. \textbf{B} Voltage traces for the inferred parameters, colored by their tRMSE. The observation is shown in white. \textbf{C} Median errors for seven commonly used electrophysiological features (as defined in \cref{sec:metrics}). Quartiles are marked as black bars. Fenrir is able to reproduce qualitative aspects of the data much better than RK.}
\label{fig:figure6}
\end{figure*}

We observed that for a two parameter HH model, learning the diffusion hyperparameter performed on par with the classical RK method, suggesting \citet{Tronarp_Bosch_Hennig_2022}'s approach did not provide any benefit in this setting (\cref{fig:figure5}). Moreover, the same performance could be achieved if the diffusion parameter was just fixed at a well selected value. Interestingly, the $\kappa$ that maximized the likelihood at the true parameters performed worse than when the diffusion was just fixed at a low value (\cref{fig:figure5}). Nonetheless, our method still yielded the best convergence, implying that active adaptation of $\kappa$ during optimization is imperative. 

To show that diffusion tempering worked for a variety of different model sizes and complexities, we benchmarked our algorithm against RK least-squares regression and the original method of \citet{Tronarp_Bosch_Hennig_2022} for the 1D pendulum (\cref{sec:PD_appendix}), the Lotka-Volterra equations (\cref{sec:LV_appendix}) with two and four free parameters as well as multiple HH models with one ($\text{HH}_1$) and two compartments ($\text{HH}_2$) and one to six free different parameters (\cref{sec:HH_appendix}). For 100 different initializations, we evaluated the algorithms based on trajectory mean squared error (tRMSE, \cref{def:trmse}), relative parameter mean squared error (pRMSE, \cref{def:prmse}), fraction of successfully converged runs (CONV), the number of iteration steps (ITER) and the number of correctly identified parameters ($\text{N}_\theta$). While the focus of our work was on the reliability of convergence we also show that the inclusion of a simple early stopping criterion (see \cref{sec:add_exp_details} for details) makes our algorithm competitive on cost.

We observed that diffusion tempering (\textsc{ours}) yielded the best results across all of our experiments in terms of tRMSE, pRMSE and the number of successfully converged initializations (\cref{tab:benchmarks}). Moreover, with our method, we recovered the true parameters from about twice to four times as often as the default Fenrir depending on the model. Even in settings where the RK baseline was not able to converge correctly at all, as was the case for the two compartment model with six parameters, our method was correct $88\%$ of the time. Additionally, in the case of the three parameter $\text{HH}_1$ model, diffusion tempering doubled the performance of Fenrir. As expected, using a schedule came at the expense of additional optimization steps. For our selected schedule, this meant about an order of magnitude more steps compared to the original method. However, the inclusion of a simple early stopping criterion (\textsc{ours+}) made diffusion tempering cost competitive without loss of performance for the subset of problems we tested. This can be attributed to the fact that convergence with a high precision is unnecessary during early tempering stages and that most trajectories converged far before the end of the schedule (\cref{fig:appendix_figure4a}). Combined with steeper schedules (\cref{sec:schedule_comparison} this would bring the costs in line with the original Fenrir method while retaining superior performance.

While for the $\text{HH}_1$ model with 6 parameters all methods seemingly performed on par with respect to mean pRMSE and tRMSE, no algorithm manages to converged to the true parameters reliably. However, this was a very complicated model, which is why we looked at the parameter estimates in detail. First we looked at the distributions of tRMSEs. 

We noticed that the tRMSEs obtained via diffusion tempering were much more uniformly distributed than those from RK, while being in the same range (\cref{fig:figure6}A). However, not only was the fraction of parameter estimates which yielded steady-state solutions for RK extremely high (\cref{fig:figure6}B) -- for many applications these are discarded immediately \cite{Prinz_Billimoria_Marder_2003, Rossant_Goodman_Fontaine_Platkiewicz_Magnusson_Brette_2011} -- but they often displayed very different spike frequencies from the observed data. This was not the case for our method for which the solutions come qualitatively much closer to that of the observed data. For practitioners interested in parameter estimation for this model, it mainly matters whether a set of commonly used eletrophysiological summary statistics is reproduced by the fitted model (for their definitions see \cref{sec:summary_stats}). Here we found that our method deviated much less from the observed behavior, suggesting that the parameters capture essential aspects of the data much better than the classical baseline (\cref{fig:figure6}C). Both methods also performed much better than when the diffusion was learned (\cref{fig:appendix_figure6}).

\section{Discussion}
Gradient descent has the potential to efficiently fit ODEs to data, but non-linear differential equations can exhibit many local minima, leading to poor convergence and extreme sensitivity to initial conditions. Here, we used probabilistic integrators to improve the convergence of gradient descent based parameter fitting of ODEs. We developed diffusion tempering, a method that successively solves several parameter search tasks. We demonstrated that our method dramatically improves the reliability of parameter optimization across several popular ODEs, and we showed that our method enables gradient descent based parameter estimation on a six-parameter Hodgkin--Huxley model, which had been unattainable with existing gradient-based methods.

\subsection{Limitations}
Of the compared methods diffusion tempering was the most expensive in terms of runtime, which we attribute to two aspects: the runtime of the ODE solver, and the sequential solving of multiple optimization problems. The probabilistic ODE solvers used by our method are semi-implicit and A-stable \cite{tronarp18}. While this allows them to perform well on stiff ODEs, this comes at the cost of cubic scaling with the ODE dimension, shared by traditional implicit Runge--Kutta methods \cite{Hairer_Wanner_1999}. Using explicit probabilistic ODE solvers, which scale linearly with the ODE dimension \citep{Krämer_Bosch_Schmidt_Hennig_2022}, would improve the runtime of diffusion tempering on larger systems. Additionally, the overall runtime of diffusion tempering is strongly dependent on the iterative optimization procedure that is used. This includes the tempering curve as well as the optimization itself. However, we did show that early stopping and steeper schedules, together, can bring down the computational cost to much closer to that of the other methods, and we suspect better tempering schemes can improve this even further. We also note that in our work we assumed the magnitude of the observation noise $R$ to be known, which in practice is often not the case and which can be taken into account by Fenrir as well.

As the scope of this paper was gradient descent based parameter estimation in ODEs, we did not include other commonly used approaches such as genetic algorithms \cite{Hazelden_Liu_Shlizerman_Shea-Brown_2023} in our experimental evaluation. Furthermore, we only looked at recovering parameters for a known form of dynamics. While optimizing parameters of NODEs \cite{Chen_Rubanova_Bettencourt_Duvenaud_2018} presents an interesting problem set, we see a few potential challenges before our method can be applied here. Firstly, computing gradients efficiently would require the use of adjoints \cite{Chen_Rubanova_Bettencourt_Duvenaud_2018}, which, to our knowledge, has not yet been derived in the probabilistic numerical framework. Additionally, the $O(d^3)$ scaling of our method in the ODE dimension would limit the number of latent dimensions that can be modeled.

\subsection{Future Work}
While we already presented ways to reduce the cost of diffusion tempering using simple early stopping criteria or different schedules, more elaborate and potentially problem aware tempering schemes could further increase the efficiency and effectiveness of our method. Additionally, in our work we assumed the observation noise to be known, hence further work is needed to study the effect of also learning or scheduling this parameter, which could have a similarly regularizing effect on the optimization.

\section*{Authorship Contribution}
Conceptualization: PB, PH, JHM, JB, NB; Methodology: JB, NB; Software: JB, NB; Investigation: JB, NB, MD; Formal Analysis: JB; Visualization: JB; Supervision: PB, PH, JHM; Writing -- Original Draft: JB, NB; Writing -- Review \& Editing: PB, NB, MD, PH, JHM, KLK; Funding Acquisition: PB, PH, JHM

\section*{Acknowledgments}
We thank the members of the Probabilistic Inference in Mechanistic Models (PIMMS) Network project of the Cluster of Excellence ``Machine Learning --- New Perspectives for Science'' for discussion. The work was funded by the German Science Foundation (EXC 2064 ``Machine Learning --- New Perspectives for Science'', 2064/1 – 390727645), the Hertie Foundation,  the European Union (ERC, ``NextMechMod'', ref. 101039115, ``ANUBIS'', ref. 101123955, ``DeepCoMechTome'', ref. 101089288), as well as the German Federal Ministry of Education and Research (BMBF) through the Tübingen AI Center (01IS18039A). Views and opinions expressed are however those of the authors only and do not necessarily reflect those of the European Union or the European Research Council Executive Agency. Neither the European Union nor the granting authority can be held responsible for them. PH and NB are supported by funds of the Ministry of Science, Research and Arts of the State of Baden-Württemberg. 
We also thank the International Max Planck Research School for Intelligent Systems (IMPRS-IS) for supporting JB, NB, MD and KLK.

\section*{Impact Statement}
This paper contributes to the advancement of the field of machine learning. It aligns with the ongoing progress in the broader landscape of machine learning research. Compared to other methods in the field our work does not introduce unique and distinct societal challenges that warrant explicit discussion.

\bibliography{references}
\bibliographystyle{icml2024}

\newpage
\appendix
\setcounter{figure}{0}
\renewcommand{\thefigure}{A\arabic{figure}}
\setcounter{section}{0}
\renewcommand{\thesection}{A\arabic{section}}

\onecolumn
\section*{\LARGE Appendix}

\section{Implementation}\label{sec:implementation}
We wrote all of our experiments in the Julia programming language \cite{Bezanson_Edelman_Karpinski_Shah_2017}. For the implementation of probabilistic ODE filters we relied on the original code for Fenrir \cite{Tronarp_Bosch_Hennig_2022} and \href{https://github.com/nathanaelbosch/ProbNumDiffEq.jl}{ProbNumDiffEq.jl}. To compute the trajectories for the observations as well as for least-squares regression with RK, the \texttt{RadauIIA5} \citep{Hairer_Wanner_1999} solver is used, with adaptive step-size selection. The solver was provided by DifferentialEquations.jl \cite{Rackauckas_Nie_2017}. For the optimization we used the implementation of L-BFGS from Optim.jl \cite{Mogensen_Riseth_2018}. This is a comparable setup to that of \citet{Tronarp_Bosch_Hennig_2022}.

All computations were done on an internal cluster running Intel(R) Xeon(R) Gold 6226R CPUs @ 2.90GHz. All code and data are publicly available on \href{https://github.com/berenslab/DiffusionTempering}{GitHub} and \href{https://zenodo.org/records/11244700}{Zenodo}.

\section{Additional experimental details}\label{sec:add_exp_details}
To generate the data used as the observations we solved the IVP with the fully-implicit Runge--Kutta \texttt{RadauIIA5} \citep{Hairer_Wanner_1999} method with both absolute and relative tolerances of $10^{-14}$ on an equi-spaced time grid $t_i \in \mathbb{T} = \{0.0, 0.01, ..., T\}$ with $dt=0.01$. The $T$s differed for each model and can be found in \cref{sec:models_appendix}. Uniform Gaussian noise of variance $\sigma^2=0.1$ was then added to the data.

To evaluate the different methods we conducted all experiments for 100 different initializations.

As the probabilistic ODE solver for the IVP we used a first order linearization of the vector field and a 3-times integrated integrated Wiener process prior.

During diffusion tempering the optimizer often converged very close to the parameter bounds. This meant that the initial values for the next optimization problem would lie right on top of the bounding box, which would throw an error. We mitigate this, by giving a tiny nudge to any initial values that end up right on top of the parameter bounds before starting the optimization.

In the experiments marked with \textsc{ours+} in \cref{tab:benchmarks} we added a simple early stopping criterion to \cref{alg:diffusion_tempering} to reduce the computational cost of our algorithm. This could prematurely stop the optimization (OPT) during all but the last scheduled value of $\kappa$, if the absolute change in the Fenrir log likelihood $\widehat{\mathcal{M}}_{PN}$ across 3 consecutive updates was below $0.1$. Since last tempering step was never interrupted, this ensured the final optimization was performed to the same tolerances as the other methods. 

\section{Metrics}\label{sec:metrics}
To compare the quality of the parameter estimates we use several metrics. Firstly, we compare using the trajectory root mean squared error (RMSE)
\begin{definition}[Trajectory RMSE]
  \label{def:trmse}
  Let \(\hat{\theta}\) be the parameters estimated by an inference algorithm, and let \(\mathbb{T}_\mathsf{D}\) be the set of measurement nodes.
  Then, let \(\hat{y}(t)\), \(t \in \mathbb{T}_\mathsf{D}\), be the estimated system trajectory, computed by numerically integrating the ODE with initial values and parameters as given by the estimated \(\hat{\theta}\).
  The trajectory RMSE (tRMSE) is then defined as
  \begin{equation}
    \text{tRMSE} := \sqrt{\frac{1}{\left| \mathbb{T}_\mathsf{D} \right|} \sum_{t \in \mathbb{T}_\mathsf{D}} \left\| \hat{y}(t) - y(t) \right\|_2^2}.
  \end{equation}
\end{definition}
which evaluates the quality of the results in the trajectory space and which is also used in the original work\cite{Tronarp_Bosch_Hennig_2022}. However, while this is useful to get an idea of how closely the observed data was reproduced it does not necessarily allow to deduce how well the parameters match those that have generated the original data. This is especially a problem for oscillatory systems for which the tRMSE of the zero function is often better than for a solution which is slightly phase shifted, which arguably is often a better fit. We therefore evaluate the quality of the inferred trajectories also in parameter space. 

\begin{definition}[Relative Parameter RMSE]
  \label{def:prmse}
  Let \(\hat{\theta}\) be the parameters estimated by an inference algorithm, and let \(\theta^*\) be the known, true parameters which are to be inferred. Let $D$ be the dimensionality of the parameter space.
  Then the relative parameter RMSE (pRMSE) \(d_\theta\) is defined as 
  \begin{equation}
    \text{pRMSE} := \sqrt{\frac{1}{D}\sum_{i=1}^D \left\|(\hat{\theta}_i - \theta_i^*)/\theta_i^*\right\|_2^2}.
  \end{equation}
\end{definition}

This metric is also used to determine whether an optimization converged successfully or not. If the final parameter estimate $\hat{\theta}$ lies within a 5\% ball of the true parameters, i.e. $pRMSE < 0.05$, then we consider it to have correctly converged. 

\subsection{Electrophysiological Summary Features}\label{sec:summary_stats}
It is difficult to evaluate the quality of parameter estimates for higher parameterised HH models, since a good tRMSE does not necessarily mean characteristic features of the data are being reproduced. Hence to score the quality of parameter estimates, we also define seven commonly used electrophysiological summary features. These reflect measures that electrophysiologists regularly use to characterize and quantify the qualitative behavior of recorded neurons \cite{Scala_et_al_2020}. This reduces the high dimensional model output to a meaningful set of numbers that emphasize specific aspects of the data and are easier to interpret.

\begin{definition}[Summary feature]
  \label{def:summary_feature}
  Let \(\theta\) parameterize an IVP. and let \(y(t)\), be the system trajectory, computed. Then a summary feature is defined as a function
  \begin{equation}
    \text{$x$} := s(y(t)).
  \end{equation}
   which reduces the system trajectory to a single number \(x\).
\end{definition}

The following summary statistics are computed from the trajectory.
\begin{table}[h]
    \centering
    \caption{Summary features to quantify the behavior of different voltage traces.}
    \vskip 0.15in
    \begin{tabular}{|l|l|}
        \hline
         Feature &  Definition\\
         \hline
         AP peak & Maximum voltage of an action potential (AP). \\
         AP trough & Minimum voltage of the trough / hyperpolarization that follows the AP. \\
         AP amp & Difference of peak $V_{peak}$ and trough voltage $V_{trough}$.\\
         ISI & The time between two consecutive APs.\\
         Resting V & Average of $V(9\,ms< t < 10\,ms)$, i.e. 1 ms before the stimulus.\\
         Latency & Time between stimulus onset and reaching the first peak voltage. \\
         \#APs & Number of APs.\\
         \hline
    \end{tabular}
    \label{tab:summary_features}
\end{table}

For AP peak, AP trough, AP amp and ISI we compute the mean. To compare to the observation we then take the absolute value of the relative differences between the observed $x^*$ and the estimated $\hat{x}$ summary features. If no spikes are detected, spike dependent features will set to NaN.

\section{Optimization}\label{sec:optimization}
For optimization we use L-BFGS for both Fenrir and RK \cite{Nocedal_Wright_2006}. This is also the optimizer used in the original Fenrir paper \cite{Tronarp_Bosch_Hennig_2022}. To speed up convergence we also use backtracking line search \cite{Armijo_1966}. To determine convergence we used an absolute tolerance of $10^{-9}$ and a relative tolerance of $10^{-6}$, which would give an accuracy of about $10^{-4}$ to $10^{-5}$ since the total solution accuracy is roughly 1-2 digits less than the relative tolerances (see \href{https://docs.sciml.ai/DiffEqDocs/stable/basics/faq/}{SciML Docs}).

The optimization was performed with box-constraints, such that the optimization would stay within the limits provided by the uniform prior distribution with upper and lower bounds $\theta_{ub}, \theta_{lb}$. All initial parameters were also drawn from this distribution, with initial values set to $u(t_0)$. Observation noise is always fixed to the true $\sigma^2 = 0.1$.

All optimized parameters $\theta$ were transformed to ranges of 0 to 1 according to:

\begin{align}
    \tilde{\theta} = \frac{\theta - \theta_{lb}}{\theta_{ub} - \theta_{lb}}, \label{eq:param_transform}
\end{align}

where $\theta_{lb}$ and $\theta_{ub}$ are the lower and upper bounds on the parameter.
Additionally $\kappa$ was first transformed to log-space, before putting it through \cref{eq:param_transform}. This ensures that the gradients for each parameter are roughly comparable.

\section{Models}\label{sec:models_appendix}

\subsection{Pendulum}\label{sec:PD_appendix}
The first ODE we consider is that of a pendulum (\cref{eq:pendulum}). Although this is a fairly simple system, parameter estimation can be challenging due to local optima. The ODE is a 2nd order equation $\frac{d^2\phi}{dt^2} = -\frac{g}{l}sin(\phi)$, which we reformulate as a system of first order equations.
\begin{subequations}\label{eq:pendulum}
\begin{align}
\frac{d\phi_1}{dt} &= \phi_2\\
\frac{d\phi_2}{dt} &= -\frac{g}{l}sin(\phi_1)
\end{align}
\end{subequations}
where $g=9.81\,m/s^2$ and $\varphi(0)=\frac{\pi}{4}$. In our experiments we chose the bounds and values for the single parameter $l$ as in \cref{tab:pd_params}.
\begin{table}[h]
    \centering
    \label{tab:pd_params}
    \caption{Parameter bounds and values that were used for our experiments.}
    \vskip 0.15in
    \begin{tabular}{|c|c|c|c|}
        \hline
         Parameter &  Lower Bound & Upper bound & observation\\
         \hline
         l & 0.1 & 10.0 & 3\\
         \hline
    \end{tabular}
\end{table}

Hence $\theta=\{l\}$. The system was integrated for $t\in[0,10]$ and only partially observed with a measurement matrix (see \cref{eq:observation_model}) of $H = [1\ 0]$.

\subsection{Lotka-Volterra}\label{sec:LV_appendix}
The Lotka–Volterra or predator and prey model, can be described by a pair of first-order nonlinear differential equations (\cref{eq:lv_sys}). It describes how the populations of two species (predator and prey) evolve when they coexist in an environment. 
\begin{subequations}\label{eq:lv_sys}
\begin{align}
    \frac{dx}{dt} &= \alpha x - \beta xy\\
    \frac{dy}{dt} &= \delta xy - \gamma y
\end{align}
\end{subequations}
We set the initial populations to $x(0)=1.0$, $y(0)=1.0$ and use the bounds and values to parameterize the model as shown in \cref{tab:lv_params}.
\begin{table}[h]
    \centering
    \label{tab:lv_params}
    \caption{Parameter bounds and values that were used for our experiments.}
    \vskip 0.15in
    \begin{tabular}{|c|c|c|c|}
         \hline
         Parameter &  Lower Bound & Upper bound & observation\\
         \hline
         $\alpha$  & $1\cdot 10^{-3}$ & 5.0 & 1.5 \\
         $\beta$  & $1\cdot 10^{-3}$ & 5.0 & 1.0 \\
         $\gamma$  & $1\cdot 10^{-3}$ & 5.0 & 3.0 \\
         $\delta$  & $1\cdot 10^{-3}$ & 5.0 & 1.0 \\
         \hline
    \end{tabular}
\end{table}
Hence $\theta=\{\alpha, \beta \}$ for the 2 parameter and $\theta=\{\alpha, \beta, \gamma, \delta \}$ for the 4 parameter case. The system was integrated for $t\in[0,20]$ and only partially observed with a measurement matrix (\cref{eq:observation_model}) of $H = [0\ 1]$, such that only the prey species is observed. 

\subsection{The Hodgkin Huxley Model}\label{sec:HH_appendix}
Here we outline the specific versions of the HH model that were used in our experiments. We consider sodium, potassium and leak currents to generate spiking, a slow non-inactivating $K^+$ current to enable spike-frequency adaptation, and a high-threshold $Ca^{2+}$ current to generate bursting \citep{Pospischil_2008}: 

\begin{align}
     C\,A\frac{dV_t}{dt} &= I_t + \bar{g}_{\Nat}m^3h(E_{\Nat}-V_t) \nonumber\\
     &+ \bar{g}_{\Kt} n^4(E_{\Kt}-V_t) + \bar{g}_{\leakt}(E_{\leakt}-V_t) \label{eq:dVm/dt}\\
     &+ \bar{g}_{\Mt} p(E_{\Kt}-V_t) + \bar{g}_{\Lt}q^2r(E_{\Cat}-V_t)\nonumber\\
\end{align}

$\bar{g}_i,\ i \in \{\Nat,\Kt,\leakt,\Mt,\Lt\}$ are the maximum conductances of the sodium, potassium, leak, adaptive potassium and calcium ion channels, respectively. $E_i$ are the associated reversal potentials. $I_t$ denotes the current per unit area, $C$ the membrane capacitance, $A$ the compartment area and $n, m, h, q, r$ and $p$ represent the fraction of independent gates in the open state, based on \citet{Hodgkin_Huxley_1952}. 

The dynamics of the gates can be expressed in general form as \cref{eq:dz/dt,eq:dp/dt}, where $\alpha_z(V_t)$ and $\beta_z(V_t)$ are rate constants for each of the gating variables $z \in \{m,n,h,q,r\}$ and $p$. 

\begin{subequations}
\begin{align}
         \frac{dz_t}{dt} &= (\alpha_z(V_t)\,(1-z_t) - \beta_z(V_t) z_t) \label{eq:dz/dt}\\
    \frac{dp_t}{dt} &= ((p_{\infty}(V_t) - p_t)/\tau_p(V_t))\label{eq:dp/dt}
\end{align}
\end{subequations}

They model the kinetics of different channel proteins. The parameters for  $\alpha_z(V_t)$, $\beta_z(V_t)$, $p_{\infty}(V_t)$ and $\tau_p(V_t, \tau_{max})$ are taken from \citet{Pospischil_2008}. The detailed equations are also provided in \cref{eq:hh_rates}.
\begin{subequations}\label{eq:hh_rates}
\begin{align}
\alpha_m(V) &= -0.32 \frac{V - V_T - 13}{e^{-(V - V_T - 13) / 4} - 1},\quad \beta_m(V) = 0.28 \frac{V - V_T - 40}{e^{(V - V_T - 40) / 5} - 1} \\
\nonumber\\
\alpha_n(V) &= -0.032 \frac{(V - V_T - 15)}{e^{-(V - V_T - 15) / 5} - 1}, \quad \beta_n(V) = 0.5 e^{-(V - V_T - 10) / 40} \\
\nonumber\\
\alpha_h(V) &= 0.128 e^{-(V - V_T - 17) / 18}, \quad
\beta_h(V) = \frac{4}{e^{-(V - V_T - 40) / 5}+1} \\
\nonumber\\
\alpha_q(V) &= 0.055 \frac{-27 - V}{e^{(-27 - V) / 3.8} - 1}, \quad
\beta_q(V) = 0.94 e^{(-75 - V) / 17} \\
\nonumber\\
\alpha_r(V) &= 0.000457 e^{(-13 - V) / 50}, \quad
\beta_r(V) = 0.0065 / (e^{(-15 - V) / 28} + 1) \\
\nonumber\\
\tau_p(V) &= \frac{\tau_{\maxt}}{3.3 e^{(V + 35) / 20} + e^{-(V + 35) / 20}}
\end{align}
\end{subequations}
The parameter values and bounds we used in our experiments for the HH model can be found in \cref{tab:HH_params}.

\begin{table}[h]
    \centering
    \caption{Parameter bounds and values that were used for our experiments.}
    \vskip 0.15in
    \begin{tabular}[h]{|c|c|c|c|}
        \hline
        Parameter & Lower bound & Upper bound & Observation\\
         \hline
        $C$ ($\mu F/cm^2$) & 0.4 & 3  & 1\\
        $A$ ($cm^2$) & $1.9\cdot 10^{-5}$ & $30.2\cdot 10^{-5}$ & $8.3\cdot 10^{-5}$\\
        $g_{\Nat}$ (mS) & 0.5 & 80  & 25\\
        $g_{\Kt}$ (mS) & $1\cdot 10^{-4}$ & 15  & 7\\
        $E_{\Nat}$ (mV) & 50 & 100 & 53\\ 
        $E_{\Kt}$ (mV) & 110 & -70 & -107\\
        $g_{\leakt}$ (mS) & $1\cdot 10^{-4}$& 0.8 & 0.1\\
        $E_{\leakt}$ (mV) & -110 & -50 & -70\\
        $V_T$ (mV) & 90 & -40 & -60\\
        $g_{\Mt}$ (mS) & $1\cdot 10^{-5}$ & 0.6 & 0.01\\
        $E_{\Cat}$ (mV) & 100 & 150 & 120\\
        $g_{\Lt}$ (mS) & $1\cdot 10^{-4}$ & 0.6 & 0.01\\
        $\tau_{\maxt}$ (s) & 50 & $3\cdot 10^3$ & $4\cdot 10^3$\\
        \hline
    \end{tabular}
    \label{tab:HH_params}
\end{table}

$A=\frac{\tau}{C R_{in}}$, $C$ the membrane capacitance, $R_{in}$ the input resistance. $\tau$ is the membrane time constant, $\tau_{max}$ the time constant of the slow $K^+$ current and $V_T$ the threshold voltage.\\

The default parameters are chosen according to \url{http://help.brain-map.org/download/attachments/8323525/BiophysModelPeri.pdf}. With the parameters for the cell area $A$ being taken from in vitro recordings from the mouse cortex from the \href{http://celltypes.brain-map.org/data}{Allen Cell Type Database} the cell ID (509881736). This closely follows the experiments from \cite{Goncalves_2020}.

For the 1, 2 and 3 parameter models we removed the equations for $I_M$, and $I_L$ to make it easier to simulate. This is the same as considering $g_M$ and $g_L$ to be 0. Furthermore, for the 6 and 8 parameter model the parameters where adjusted to $gleak = 0.05\,mS$ and $\tau_{max} = 1000\,s$. The initial voltage was set to $V(0)=-70\,mV$, from which the initial gating variables where computed according to \cref{eq:hh_gates_initial_values}, where $z(0)=z_\infty(V(0))$ for $z\in\{m,n,h,p,q,r\}$. The system was integrated for $t\in[0,100]$ms and only the voltage was observed, which yields a measurement matrix of $H=[1\ 0\ 0\ 0\ 0\ 0\ 0]$. For the full system and $H=[1\ 0\ 0\ 0]$ with $I_M$ and $I_L$ removed.

\begin{subequations}\label{eq:hh_gates_initial_values}
\begin{align}
m_{\infty}(V) &= 1 / (1 + \beta_m(V) / \alpha_m(V)) \\
n_{\infty}(V) &= 1 / (1 + \beta_n(V) / \alpha_n(V)) \\
h_{\infty}(V) &= 1 / (1 + \beta_h(V) / \alpha_h(V)) \\
p_{\infty}(V) &= 1 / (1 + e^{-(V + 35) / 10}) \\
q_{\infty}(V) &= 1 / (1 + \beta_q(V) / \alpha_q(V)) \\
r_{\infty}(V) &= 1 / (1 + \beta_r(V) / \alpha_r(V)) \\
\end{align}
\end{subequations}

Depending on the model, the following parameter sets were optimized.
\begin{table}[h]
    \centering
    \caption{Parameter sets for the different single compartment models, which we denote with $\text{HH}_1$.}
    \vskip 0.15in
    \begin{tabular}[h]{|l|l|}
        \hline
        \#parameters & $\theta$\\
         \hline
        1 & $\{g_{\Nat}\}$\\
        2 & $\{g_{\Nat},g_{\Kt}\}$\\
        3 & $\{g_{\Nat},g_{\Kt}, g_{\leakt}\}$\\
        6 & $\{g_{\Nat},g_{\Kt}, g_{\Nat}, g_{\Mt}, V_T, g_{\Lt}\}$\\
        \hline
    \end{tabular}
    \label{tab:HH_param_sets}
\end{table}

\subsubsection{Multicompartment Models}\label{sec:HHmulti_appendix}
To model neurons with a complex morphological structure the neuron is split into discrete compartments that are small enough that the variation of the membrane potential across it is negligible. Then the continuous membrane potential $V$ can be approximated as by a sum of local compartment potentials. For a non-branching cable, this can be expressed as
\begin{align}
C A_i \frac{d V_i}{d t}=-I_{ionic, i}+I_i+g_{i, i+1}\left(V_{i+1}-V_i\right)+g_{i, i-1}\left(V_{i-1}-V_i\right),\label{eq:cable}
\end{align}
with subscript $i$ indexing the compartments, the sum of the ionic currents across the cells membrane $I_{ionic}$, the external current $I$ and coupling coefficients $g$. Each membrane potential $V_i$ satisfies an equation similar to \cref{eq:dVm/dt} except that each compartments couples to two neighbors (unless at the ends of the cable). For more context see \cite{Dayan_Abbott_2001} Chapter 6. In our experiments we used coupling coefficients of $g = 1$ and compartment areas of $A_i=\frac{A}{N}, \quad i\in \{1,...,N\}$, for $N$ compartments, such that the total area would sum to the same $A$ as in \cref{tab:HH_params}. Furthermore, we stimulated only the first compartment and considered all compartments to be observed (voltage only). The parameters used for the 4 and 6 parameter models where modified from \cref{tab:HH_params}. For the 4 parameter model the 2nd compartment was adjusted to $g_{Na} = 20\,mS$ and $g_{K} = 10\,mS$. For the 6 parameter model, the first $g_{leak} = 0.09\,mS$ and the 2nd compartment was changed to $g_{Na} = 20\,mS$, $g_{K} = 10\,mS$ and $g_{leak} = 0.11\,mS$. 

Depending on the number of compartments $N$ this yields parameter sets of the form $\theta=\{\theta_1,...,\theta_N\}$, where $\theta_i$ are analogous to the parameters of the single compartments in \cref{tab:HH_param_sets}. The measurements are taken with
\begin{equation}
  H =
\begin{bmatrix} 
    H_{1} & 0 & \dots \\
    \vdots & \ddots & \\
    0 &        & H_{N},
    \end{bmatrix}
\end{equation}
where $H_i$ are the measurement matrices for a single compartment. We denote the number of compartments that were used in an experiment with a subscript $\text{HH}_i$.

\section{Additional Details on \cref{sec:HH_results}}
Here we show all the trajectories and likelihoods for which a subset was presented in \cref{fig:figure4}. Additionally, we plotted the solutions at each of the initial and final parameters for each diffusion parameter.
\begin{figure}[h!]
\vskip 0.2in
\begin{center}
\centerline{\includegraphics[width=\columnwidth]{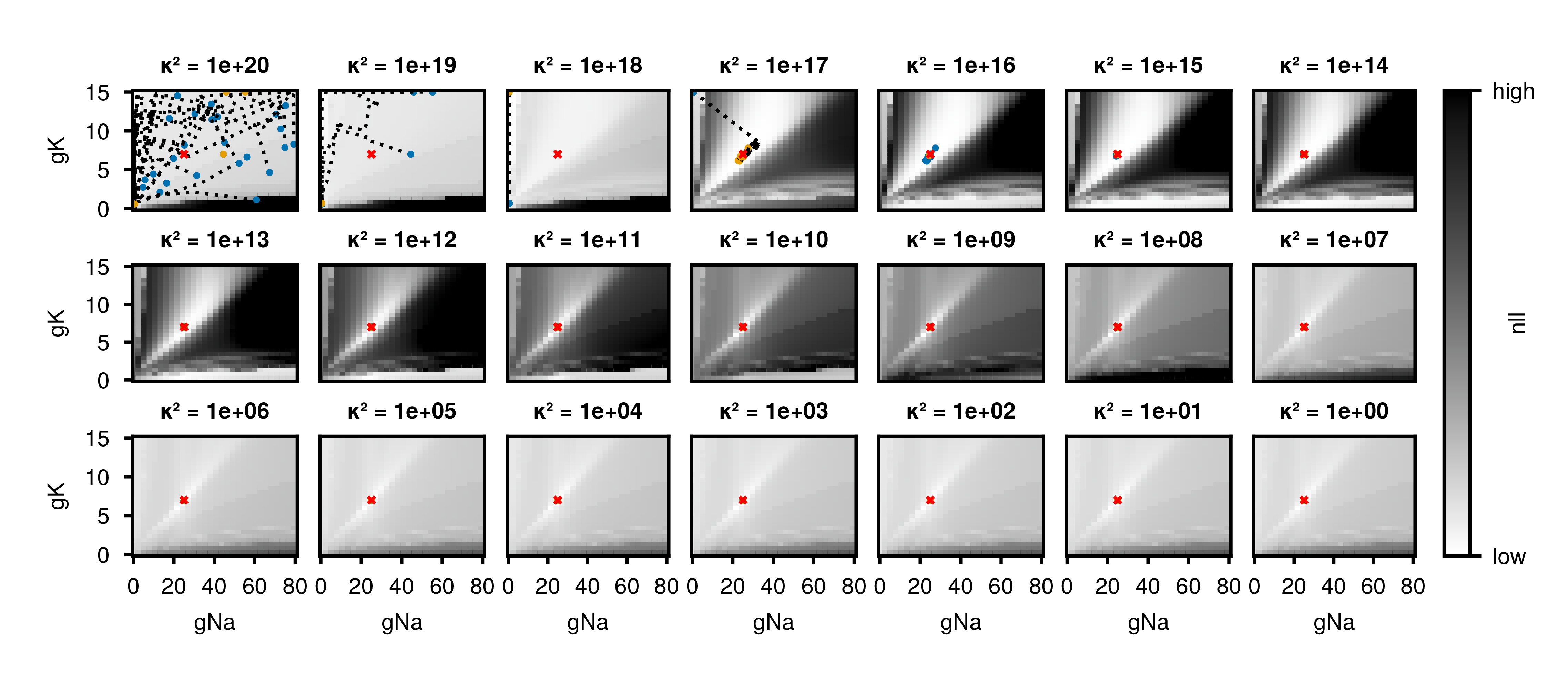}}
\caption{Parameter estimation for the sodium $g_{Na}$ and potassium $g_{K}$ conductance of a HH model. Parameter optimization during all stages of diffusion tempering for a random subset of 25 initializations. Optimization trajectories (dotted) in each likelihood landscape are shown from start (blue) to convergence (orange). Very high loss values were clipped for better visual clarity.}
\label{fig:appendix_figure4a}
\end{center}
\vskip -0.2in
\end{figure}

\begin{figure}[h]
\vskip 0.2in
\begin{center}
\centerline{\includegraphics[width=\columnwidth]{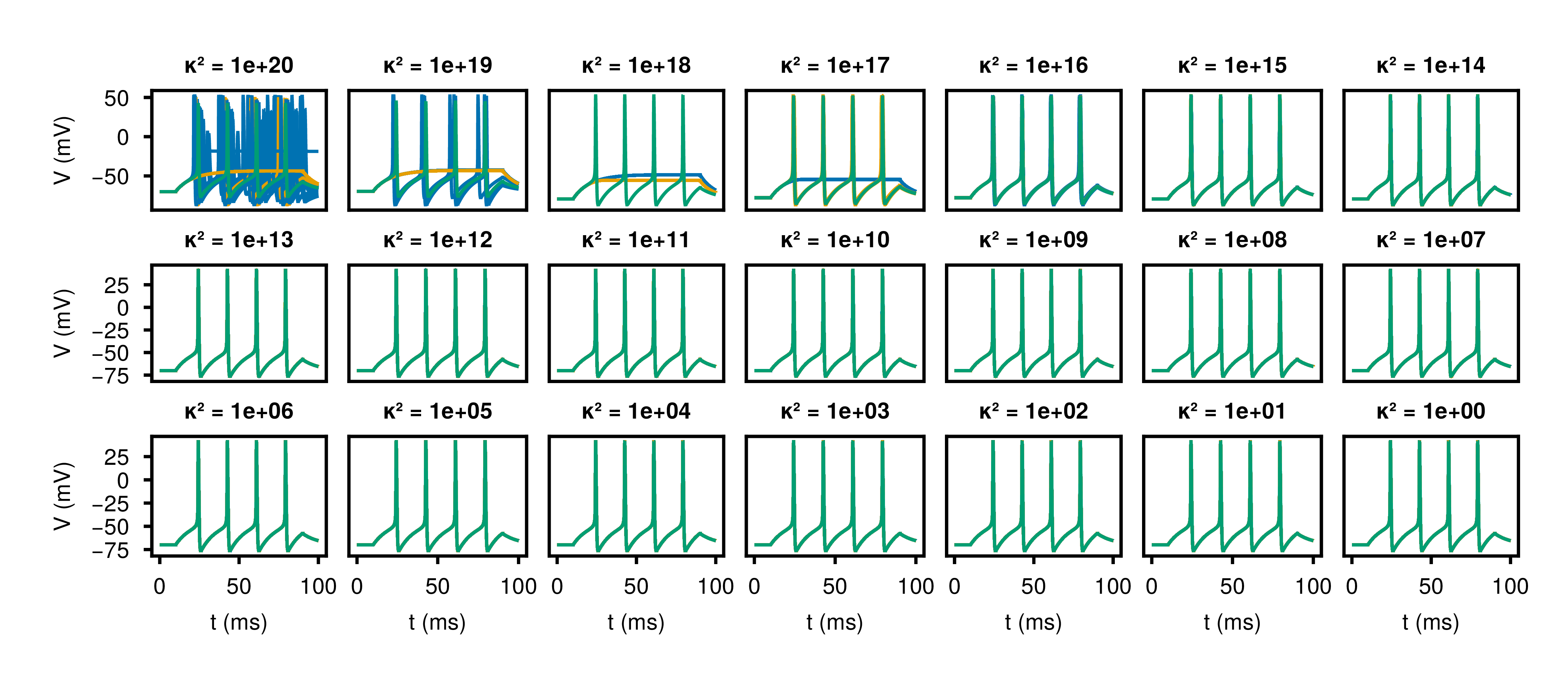}}
\caption{Parameter estimation for the sodium $g_{Na}$ and potassium $g_{K}$ conductance of a HH model. ODE solutions for the start (blue) to converged (orange) parameters at all stages of diffusion tempering for a random subset of 25 initializations.}
\label{fig:appendix_figure4b}
\end{center}
\vskip -0.2in
\end{figure}

While the likelihood is initially very flat and the uniformly initialized parameters collect on the top left edge, a basin around the global optimum appears at around $\kappa=10^{18}$ (\cref{fig:appendix_figure4a}). At $\kappa=10^{17}$ the ridge that separates the local optimum from the basin is small enough for the optimizer to cross into the basin in which the global optimum resides. With decreasing $\kappa$ the basin starts to taper more and more towards the true parameters, which becomes very distinct down to about $\kappa=10^{6}$, after which point hardly any changes to the likelihood landscape are visible.

We observed that first all runs converge to the steady state solution at $\kappa=10^{18}$, before suddenly developing spikes and converging correctly (\cref{fig:appendix_figure4b}). After reaching a diffusion parameter of $\kappa=10^{15}$, the subset of initial values has fully converged and hardly any changes are visible.

\section{Additional Details on \cref{sec:benchmark_results}}
\begin{figure}[h]
\vskip 0.2in
\begin{center}
\centerline{\includegraphics[width=\columnwidth]{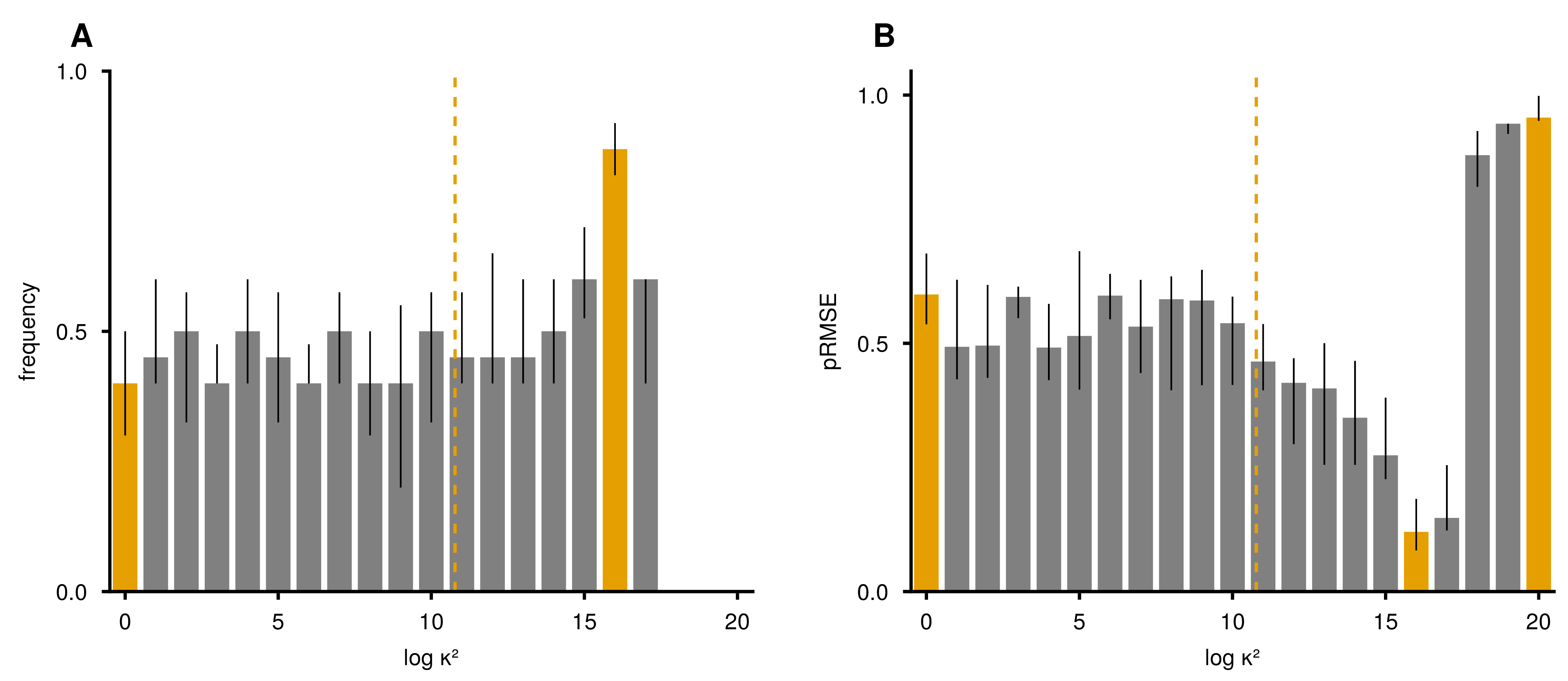}}
\caption{Convergence for a 2 parameter HH model for a grid of different fixed $\kappa$s. The parameters selected for \cref{fig:figure5} are highlighted in orange and the optimal diffusion parameter $\log \kappa = 11.6$ (dashed). Histogram shows the medians, with black bars indicating quartiles for 100 runs split into groups of 10. \textbf{A} Fraction of runs with $pRMSE<5\%$. \textbf{B} pRMSEs for each diffusion.}
\label{fig:appendix_figure5}
\end{center}
\vskip -0.2in
\end{figure}

Once the diffusion hyperparameter is set to $\log \kappa\leq 17$ most values perform similarly well in terms of recovering the true parameters about half of the time, with the exception of $\log \kappa = 16$, which does so with about 80\% reliability (\cref{fig:appendix_figure5}). Additionally, the average distance from the true parameters starts decreasing between $\log \kappa = 9$ and $\log \kappa = 17$, before sharply jumping again. The fact that the fraction of parameters that converge correctly does not change significantly until hitting $\log \kappa= 1e15$, suggests that the same set of, favorable, initial guesses converge no matter the loss. However, since the mean pRMSE starts decreasing overall, suggests that  initializations further away also end up closer to the global optimum.

Comparison of Fenrir and Diffusion tempering for a 6 parameter HH model.
\begin{figure}[h]
\vskip 0.2in
\begin{center}
\centerline{\includegraphics[width=\columnwidth]{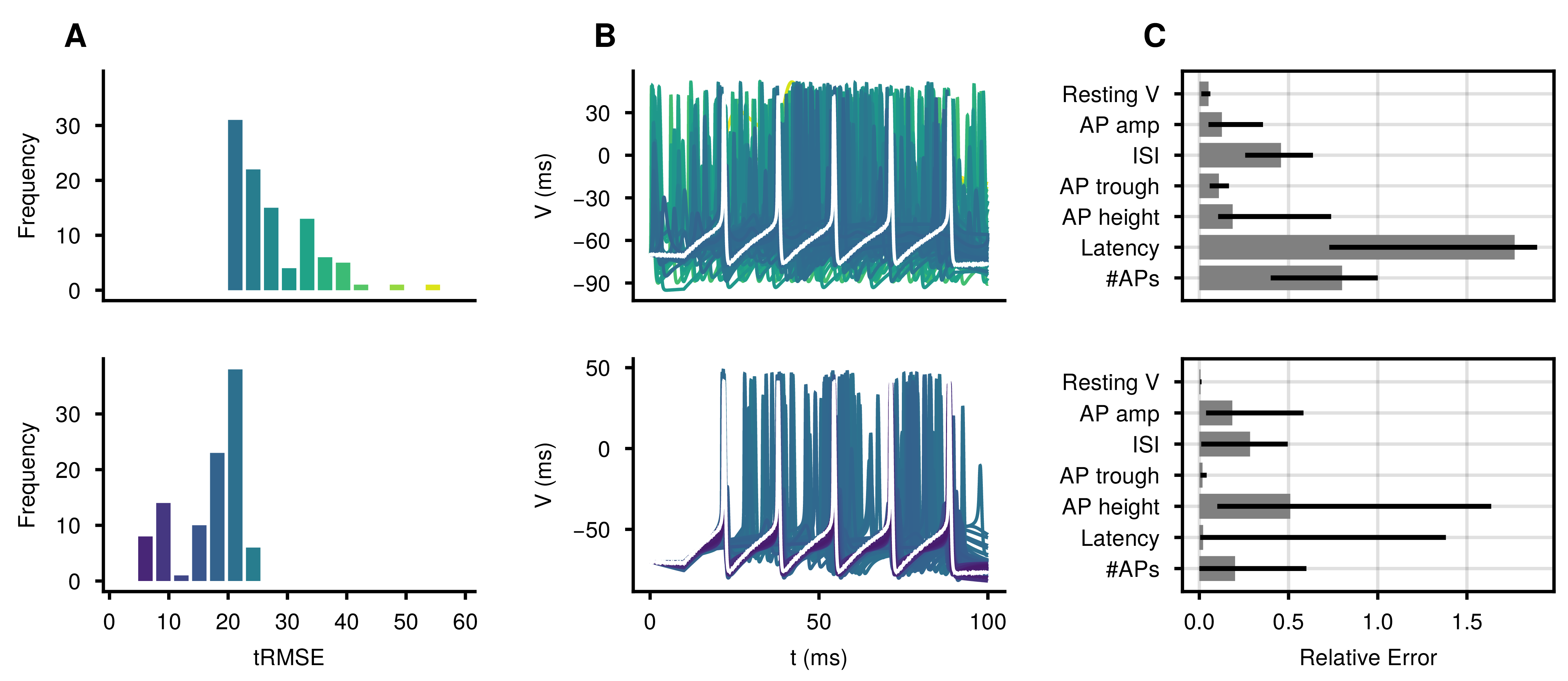}}
\caption{Parameter inference for a six parameter HH model for 100 initializations of Fenrir (top) and our method (bottom). \textbf{A} Distribution of tRMSEs. Almost all parameter estimates returned by our method have better tRMSEs than Fenrir. \textbf{B} Voltage traces for the inferred parameters, colored by their tRMSE. The observation is shown in white. The traces recovered by Fenrir seem almost random. \textbf{C} Median errors for seven commonly used electrophysiological features (as defined in \cref{sec:metrics}). Quartiles are highlighted with error bars. Our method is able to reproduce qualitative aspects of the data better than Fenrir.}
\label{fig:appendix_figure6}
\end{center}
\vskip -0.2in
\end{figure}
Diffusion tempering yields electrophysiological summary statistics much closer to those of the true observation compared to Fenrir.

\section{Additional Experiments}
\subsection{Effectiveness of different tempering rates}\label{sec:schedule_comparison}
In our work we mainly focused on the convergence reliability of diffusion tempering. For this reason we picked a tempering schedule that was very simple and reliable during our testing -- linearly reducing the $\log\kappa$ from $20$ and $0$ with a step size of 1. Despite its simplicity this already performed much better than our baselines without any further tuning. Since this schedule is very costly in the number of iterations needed (\cref{tab:benchmarks}), we also investigated the effect of different schedules, i.e. with much steeper tempering rates or exponentially decreasing $\log\kappa$ on the cost of diffusion tempering for the $\text{HH}_1$ model with 2 parameters.

\begin{figure}[h]
\vskip 0.2in
\begin{center}
\centerline{\includegraphics[width=\columnwidth]{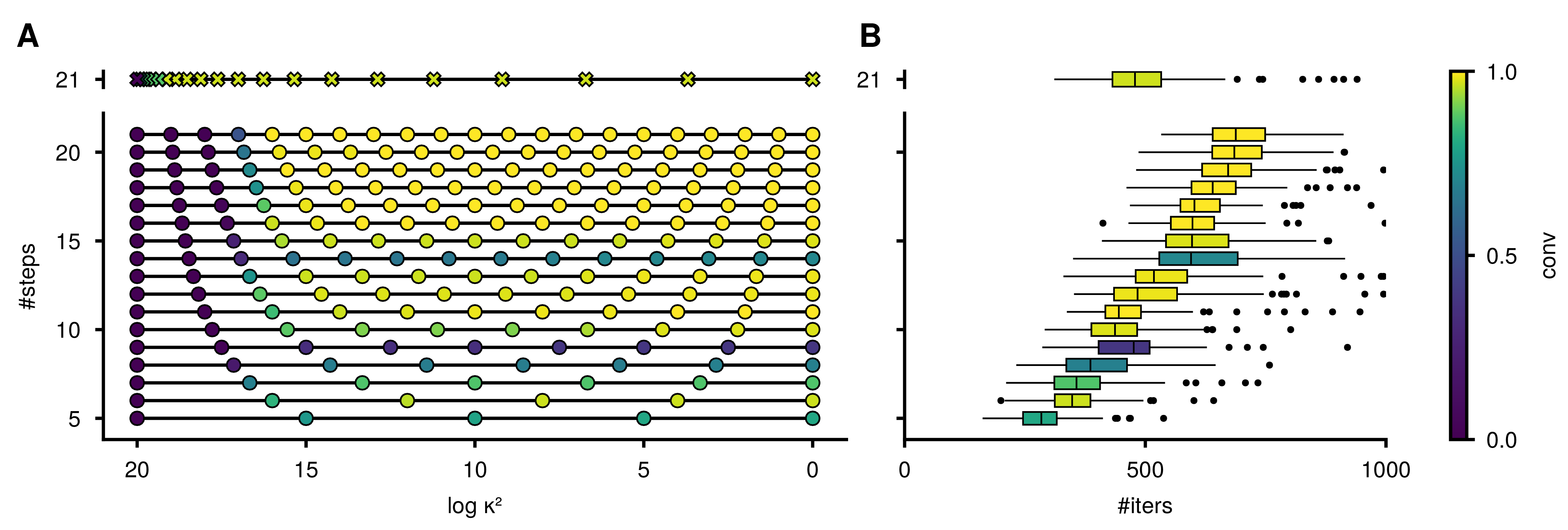}}
\caption{Comparing tempering schedules with different numbers of tempering steps between $\log \kappa=20$ and $\log \kappa=0$ for the $\text{HH}_1$ model with 2 parameters. \textbf{A} Percentage of correctly inferred parameters at each stage of tempering during optimization. Linear decrease of $\log \kappa$ (dots) vs. exponential decrease (crosses). There is a clear trade-off between the number of tempering steps and convergence, but taking more than 16 steps yields no additional benefits. \textbf{B} Cost to performance trade-off for different tempering schedules. Exponential schedules are more efficient at the same number of tempering steps.}
\label{fig:appendix_figure7}
\end{center}
\vskip -0.2in
\end{figure}

We found that much steeper, hence shorter schedules that decrease $\log\kappa$ between 20 and 0 can save about 20\% iterations, while affecting the performance not at all or very little (\cref{fig:appendix_figure7}). In particular we found that taking more than 16 tempering steps yields no additional benefits for the $\text{HH}_1$ model with 2 parameters. Additionally, similar to \cref{sec:HH_results} most of the trajectories started to converge to the true parameters around $\kappa=10^{15}$, for many of the schedules.

We also tested an exponentially decaying the $\log_{10}(\kappa)$ where at each tempering step $\kappa$ was set according to:
\begin{align}
    \mathcal{T}(i,\kappa_0) = 10^{\kappa_0 \exp(-i / \tau)},
\end{align}

with initial diffusion $\kappa_0 = 20$, decay constant $\tau = 5$ and tempering steps $i \in \{0,1,...,20\}$. We found that this exponential tempering schedule can save around 30\% of the computation cost, while only taking a small hit to performance. Combined with the early stopping criterion we implemented (\cref{sec:add_exp_details}), this brings the cost of diffusion tempering down to that of the original Fenrir method, while retaining the superior convergence.

While we only considered a small number of possible tempering schemes, our experiments show that this can already reduce the cost of our algorithm by a third. This suggests future work on more efficient tempering schemes will reduce the cost of diffusion tempering even further.

\subsection{Injecting gradient noise during optimization}
It is known that injecting noise into the gradient updates can effectively regularize the optimization procedure in challenging loss landscapes \cite{Neelakantan_Vilnis_Le_Kaiser_Kurach_Sutskever_Martens_2017}. While Fenrir and hence diffusion tempering are are noise-free methods, we also investigated how adding gradient noise to RK compares to diffusion tempering for the pendulum. 

For this purpose we optimized the pendulum as described in (\cref{sec:optimization}), but Gaussian noise $\epsilon_n$ of the form $\epsilon_n \sim \mathcal{N}(0,\sigma^2e^{-n/\tau})$, with decay rate $1/\tau$ and initial variance $\sigma^2$, was added to the gradients of the log likelihood at every parameter update step $n$.

\begin{table*}[h]
    \centering
    \caption{Effect of injecting gradient noise with different scale, $\sigma^2$, and decay rates, $\tau$, during optimization of a pendulum on the convergence. While injecting gradient noise improves convergence compared to the noise-free RK baseline (first row), both Fenrir and diffusion tempering still converge more reliably.}
    \begin{sc}
\begin{tabular}{lllll}
  \hline
  $\mathbf{\sigma^2}$ & $\mathbf{\tau}$ & \textbf{conv} & \textbf{pRMSE} & \textbf{iter} \\
  \hline
  0.00 & $-$ & 0.32 & $1.42 \pm 1.09$ & $33.47 \pm 18.11$ \\
  0.01 & $-$ & 0.35 & $1.40 \pm 1.11$ & $33.79 \pm 12.84$ \\
  0.05 & $-$ & 0.34 & $1.41 \pm 1.10$ & $29.89 \pm 10.77$ \\
  0.10 & $-$ & 0.37 & $1.40 \pm 1.13$ & $27.30 \pm 11.84$ \\
  0.50 & $-$ & 0.49 & $1.14 \pm 1.13$ & $15.19 \pm 10.72$ \\
  1.00 & $-$ & 0.63 & $0.77 \pm 1.05$ & $15.52 \pm 14.12$ \\
  5.00 & $-$ & $\mathbf{0.64}$ & $0.51 \pm 0.88$ & $13.01 \pm 8.87$ \\
  10.00 & $-$ & 0.39 & $0.64 \pm 0.93$ & $10.43 \pm 7.19$ \\
  50.00 & $-$ & 0.45 & $0.49 \pm 0.83$ & $11.66 \pm 8.26$ \\
  100.00 & $-$ & 0.28 & $\mathbf{0.44 \pm 0.74}$ & $\mathbf{10.26 \pm 7.42}$ \\
  \hline
  0.00 & 5 & 0.32 & $1.42 \pm 1.09$ & $33.47 \pm 18.11$ \\
  0.01 & 5 & 0.31 & $1.45 \pm 1.08$ & $35.94 \pm 18.58$ \\
  0.05 & 5 & 0.33 & $1.45 \pm 1.10$ & $32.58 \pm 13.39$ \\
  0.10 & 5 & 0.35 & $1.40 \pm 1.11$ & $33.65 \pm 12.85$ \\
  0.50 & 5 & 0.43 & $1.24 \pm 1.13$ & $22.23 \pm 14.93$ \\
  1.00 & 5 & 0.53 & $0.98 \pm 1.08$ & $22.68 \pm 18.51$ \\
  5.00 & 5 & $\mathbf{0.59}$ & $0.67 \pm 0.99$ & $15.23 \pm 13.33$ \\
  10.00 & 5 & 0.45 & $0.51 \pm 0.84$ & $13.11 \pm 13.49$ \\
  50.00 & 5 & 0.40 & $\mathbf{0.51 \pm 0.79}$ & $\mathbf{11.36 \pm 11.60}$ \\
  100.00 & 5 & 0.41 & $0.58 \pm 0.86$ & $12.54 \pm 14.20$ \\
  \hline
  0.00 & 10 & 0.32 & $1.42 \pm 1.09$ & $33.47 \pm 18.11$ \\
  0.01 & 10 & 0.33 & $1.43 \pm 1.10$ & $33.42 \pm 12.90$ \\
  0.05 & 10 & 0.32 & $1.44 \pm 1.09$ & $34.21 \pm 11.00$ \\
  0.10 & 10 & 0.34 & $1.39 \pm 1.11$ & $33.40 \pm 12.70$ \\
  0.50 & 10 & 0.46 & $1.22 \pm 1.14$ & $18.89 \pm 13.20$ \\
  1.00 & 10 & 0.50 & $1.08 \pm 1.10$ & $18.58 \pm 14.95$ \\
  5.00 & 10 & $\mathbf{0.61}$ & $0.58 \pm 0.94$ & $14.28 \pm 12.63$ \\
  10.00 & 10 & 0.58 & $0.55 \pm 0.90$ & $12.95 \pm 10.90$ \\
  50.00 & 10 & 0.32 & $0.41 \pm 0.71$ & $\mathbf{9.96 \pm 8.40}$ \\
  100.00 & 10 & 0.37 & $\mathbf{0.40 \pm 0.72}$ & $10.92 \pm 8.47$ \\
  \hline
      \multicolumn{2}{c}{Fenrir} & 0.75 & $0.20 \pm 0.37$ & $110.04 \pm 61.70$ \\
    \multicolumn{2}{c}{ours} & \textbf{1.00} & $\mathbf{0.00 \pm 0.00}$ & $696.22 \pm 78.10$ \\
\end{tabular}
    \label{tab:noisy_gradients}
    \end{sc}
\end{table*}

The addition of gradient noise with varying scales and decay rates improved the performance of the noise-free RK variant almost across the board, even doubling the convergence rates in some instances (\cref{tab:noisy_gradients}). Adding noise with a variance at around 5 performed the best reaching a convergence rate of $0.64$ for non decaying noise levels. While this improves upon the noise-free RK baseline, this still falls short of the convergence rates that both Fenrir and diffusion tempering achieve however, with 0.75 and 1.00 respectively. 
\end{document}